\documentclass[final,1p,times,authoryear]{elsarticle}

\usepackage{amssymb}

\usepackage{amsmath}

\usepackage{makecell}
\usepackage{float}
\usepackage{enumitem}
\usepackage{amsmath}
\usepackage{subfigure}
\usepackage{float}
\usepackage{cite}
\usepackage{svg}
\usepackage{amsmath}
\usepackage{graphics}
\usepackage{graphicx}
\usepackage{amssymb}
\usepackage{algorithm}
\usepackage{upgreek}
\usepackage{mathrsfs}
\usepackage{siunitx}
\usepackage{booktabs}
\usepackage{bm}
\usepackage{bbm}
\usepackage{makecell}
\usepackage{multirow}
\usepackage{multicol}
\usepackage{float}
\usepackage{url}
\usepackage{natbib}
\usepackage{amsfonts}
\usepackage{textcomp}
\usepackage{supertabular}
\usepackage{array}
\usepackage{hyperref}
\usepackage{amsfonts}

\graphicspath{{images/}}
\usepackage{siunitx}

\makeatletter
\def\ps@pprintTitle{%
  \let\@oddhead\@empty
  \let\@evenhead\@empty
  \let\@oddfoot\@empty
  \let\@evenfoot\@empty}
\makeatother

\begin{document}
\let\WriteBookmarks\relax
\def\floatpagepagefraction{1}
\def\textpagefraction{.001}

\begin{frontmatter}

\title{Fine-grained Hierarchical Crop Type Classification from Integrated Hyperspectral EnMAP Data and Multispectral Sentinel-2 Time Series: A Large-scale Dataset and Dual-stream Transformer Method}

\author[a]{Wenyuan Li}
\author[a]{Shunlin Liang\corref{corresponding author}}
\ead{shunlin@hku.hk}
\author[a]{Yuxiang Zhang}
\author[b]{Liqin Liu}
\author[b]{Keyan Chen}

\author[a]{Yongzhe Chen}
\author[a]{Han Ma}
\author[a]{Jianglei Xu}
\author[a]{Yichuan Ma}
\author[a]{Shikang Guan}
\author[b]{Zhenwei Shi}

\cortext[corresponding author]{Corresponding author: Shunlin Liang.}

\address[a]{Jockey Club STEM Lab of Quantitative Remote Sensing, Department of Geography, The
University of Hong Kong, Hong Kong, China}
\address[b]{Department of Aerospace Intelligent Science and Technology, School of Astronautics, Beihang University, Beijing, China}


\begin{abstract}
Fine-grained crop type classification serves as the fundamental basis for large-scale crop mapping and plays a vital role in ensuring food security. It requires simultaneous capture of both phenological dynamics (obtained from multi-temporal satellite data like Sentinel-2) and subtle spectral variations (demanding nanometer-scale spectral resolution from hyperspectral imagery). Research combining these two modalities remains scarce currently due to challenges in hyperspectral data acquisition and crop types annotation costs. To address these issues, we construct a hierarchical hyperspectral crop dataset (H$^2$Crop) by integrating 30m-resolution EnMAP hyperspectral data with Sentinel-2 time series. With over one million annotated field parcels organized in a four-tier crop taxonomy, H$^2$Crop establishes a vital benchmark for fine-grained agricultural crop classification and hyperspectral image processing. We propose a dual-stream Transformer architecture that synergistically processes these modalities. It coordinates two specialized pathways: a spectral-spatial Transformer extracts fine-grained signatures from hyperspectral EnMAP data, while a temporal Swin Transformer extracts crop growth patterns from Sentinel-2 time series. The designed hierarchical classification head with hierarchical fusion then simultaneously delivers multi-level crop type classification across all taxonomic tiers.
Experiments demonstrate that adding hyperspectral EnMAP data to Sentinel-2 time series yields a 4.2\% average F1-scores improvement (peaking at 6.3\%). Extensive comparisons also confirm our method's higher accuracy over existing deep learning approaches for crop type classification and the consistent benefits of hyperspectral data across varying temporal windows and crop change scenarios. 
Codes and dataset will be available at \url{https://github.com/flyakon/H2Crop}.
\end{abstract}


\end{frontmatter}

\section{Introduction}
\label{sec:introduction}
Fine-grained crop type classification serves as the cornerstone for global food security, agricultural policy formulation, and precision farming management. Satellite remote sensing has become a pivotal tool for large-scale agricultural monitoring due to its broad coverage and cost-effectiveness \citep{yuan2020deep,weiss2020remote,victor2024remote,liang2024advances,fang2024comprehensive}. The European Space Agency (ESA)’s Sentinel-2 satellites, with their 10-meter spatial resolutions and multispectral bands, have significantly enhanced the ability to identify spatial details in fine-grained crop classification \citep{blickensdorfer2022mapping,segarra2020remote}. However, its spectral resolution limits its capability to differentiate crops with highly similar morphological and phenological traits. 

Hyperspectral imagery (HSI), with its nanometer-level spectral resolution, enables the detection of subtle biochemical differences across crops, presenting new opportunities to distinguish crop species that are challenging for traditional multispectral  imagery (MSI) \citep{moharram2023land, guerri2024deep, delogu2023using, zhang2023sanet,liu2023diverse}. Despite HSI’s unique advantages in spectral discrimination, its limited spatiotemporal coverage (e.g.,  long revisit cycles) hinders its standalone application in large-scale dynamic monitoring. Consequently, exploring the synergistic potential of hyperspectral data and Sentinel-2 time series could pave a critical path to overcoming current accuracy bottlenecks in crop classification, though this direction remains underexplored.

The critical role of remote sensing crop type classification has driven rapid methodological advancements in this field. Phenology-based methods, as the earliest attempt, rely on carefully designed spectral indices or phenological rules to identify specific crops by leveraging their unique spectral signatures at key growth phases. Han et al. \citep{han2021nesea} developed a rule-based framework to generate annual rice distribution maps across Northeast and Southeast Asia (2017–2019). This method integrated MODIS time-series-derived spectral indices (e.g., Land Surface Water Index (LSWI) and Enhanced Vegetation Index (EVI)) with Sentinel-1-based flooding signals, effectively improving rice mapping accuracy through multi-sensor synergy. Similarly, Chen et al. \citep{liang2024mapping} employed Sentinel-1 SAR data to extract paddy rice transplanting signals, producing monsoon Asia-scale rice maps from 2018 to 2021. 

For more crops, Qiu et al. \citep{qiu2022maps} pioneered a phenology-driven algorithm using 500-m MODIS data to map China’s multiple cropping systems, incorporating pixel purity thresholds to characterize cropping intensity and spatial patterns of three staple crops (maize, rice, and wheat). However, such approaches remain heavily reliant on expert-defined thresholds and phenological assumptions, rendering them primarily effective for monoculture systems with distinct phenological signatures. Their performance significantly degrades in regions with complex crop rotations or morphologically similar species, highlighting critical scalability limitations.

To address the limitations of phenology-driven approaches, machine learning methods have emerged as an effective solution. These methods—including Random Forest (RF) \citep{breiman2001random}, Support Vector Machines (SVM) \citep{hearst1998support}, and etc.—automatically learn nonlinear relationships between satellite observations and crop types under the guidance of annotated data, thereby reducing dependency on expert-defined rules.  You et al. \citep{you202110} demonstrated this capability by extracting multi-dimensional features (spectral, temporal, and textural) from Sentinel-2 imagery, achieving high overall accuracies  for maize, soybean, and rice mapping across diverse agroecological zones. Peng et al. \citep{peng2023twenty} leveraged over 50,000 field survey samples and Time-Weighted Dynamic Time Warping (TWDTW) \citep{maus2016time,dong2020early} to reconstruct two-decade maize distribution maps in China, showcasing the scalability of machine learning in crop monitoring. Notably, van et al. \citep{van2023worldcereal} advanced global-scale agricultural cartography by deploying RF classifiers to map croplands and staple crops (wheat, maize), underscoring the paradigm-shifting potential of data-driven approaches.

While machine learning (ML) methods have been widely adopted in crop classification and mapping and achieved substantial progress, their limited model capacity and the inherent complexity of crop classification tasks restrict current applications to dominant staple crops (e.g., wheat, maize, rice). Most ML-based frameworks struggle to differentiate morphologically similar species or crops with overlapping phenological cycles, primarily due to insufficient nonlinear modeling capabilities.

Deep learning (DL) \citep{lecun2015deep}, with its superior feature extraction ability, has revolutionized pattern recognition across domains—from computer vision \citep{turkoglu2021crop,zhang2023sanet},  natural language processing \citep{guo2025deepseek,zheng2024review}  to remote sensing data understanding \citep{li2021geographical,chen2024rsprompter}. Given sufficient training data, DL models can approximate arbitrarily complex decision boundaries, offering transformative potential for fine-grained crop classification . Among DL architectures, Convolutional Neural Networks (CNNs) \citep{he2016deep,long2015fully} stand as  an effective method that employs cascaded convolutional and pooling layers to progressively extract spatial features. Although CNNs excel at processing static imagery, crop classification task demands the ability to jointly exploit temporal phenological trajectories and spatial context \citep{joshi2023remote}. To bridge this gap, 3D CNNs \citep{hara2018can, fayyaz20213d, zhou2020spatiotemporal} extend traditional architectures by incorporating temporal convolutions, enabling joint spectral-spatial-temporal feature learning.  Gallo et al. proposed a 3D CNN framework for Sentinel-2 time series processing, achieving real-time in-season crop mapping through dynamic time-series modeling \citep{gallo2023season}. 

In scenarios where ground truth is derived from field surveys, crop classification samples typically correspond to time series of individual pixels rather than image sequences. This makes Long Short-Term Memory (LSTM)  networks \citep{zhu2017deep, ghojogh2023recurrent} more suitable than Convolutional Neural Networks (CNNs) to model temporal dependencies for crop type classification.  For instance, Rußwurm et al. \citep{russwurm2023end} designed an end-to-end LSTM framework for in-season crop identification, enabling early prediction of crop types during growing cycles. When processing spatiotemporal image sequences (e.g., multi-date satellite imagery), hybrid architectures combining CNNs and LSTMs have also gained traction: CNNs extract spatial-spectral features from individual images, while LSTMs model temporal evolution. Turkoglu et al. \citep{turkoglu2021crop} exemplified this paradigm, achieving fine-grained hierarchical  crop classification through spatiotemporal feature fusion.

Building upon these advancements, Transformer-based architectures (e.g., Vision Transformers, Swin Transformers) \citep{vaswani2017attention, dosovitskiy2020image, liu2021swin, liu2022video} have recently demonstrated superior capability in jointly learning spatial-temporal representations. Their self-attention mechanisms enable global context modeling across both spatial and temporal dimensions, offering new opportunities for fine-grained crop classification. Fang et al. \citep{fang5138538generating} proposed a Transformer-based model for rice mapping in Asia, while Li et al. \citep{li5029097asiawheat} integrated CNNs with Transformers to efficiently reconstruct two-decade wheat cultivation patterns across the Asia. 

The rise of deep learning, particularly Transformer-based foundation models \citep{PIS,li2025agrifmmultisourcetemporalremote}, has unlocked unprecedented capacity to process massive-scale earth observation data. However, DL’s performance remains constrained by the scarcity of high-quality labeled data—a critical challenge in crop classification given the labor-intensive nature of field surveys. Fortunately, many European countries have recently opened access to Land Parcel Identification System (LPIS) data \citep{owen2016land}, providing parcel-level crop records that facilitate large-scale supervised learning. Leveraging these resources, researchers have constructed several benchmark datasets \citep{russwurm2019breizhcrops, weikmann2021timesen2crop,turkoglu2021crop, sykas2022sentinel, selea2023agrisen, schneider2023eurocrops} for DL-based crop classification (see Table \ref{tab:datasets}). Among them, EuroCrops \citep{schneider2023eurocrops} stands out as the most extensive dataset, harmonizing national LPIS records across EU member states while introducing hierarchical crop taxonomy.

While existing datasets and methods have significantly advanced fine-grained crop classification, their reliance on multispectral satellites like Sentinel-2 inherently limits their ability to capture subtle biochemical variations in crops, which is critical for discriminating morphologically similar species. HSI offers a transformative solution: its nanometer-level spectral resolution enables precise quantification of biochemical parameters, revealing spectral fingerprints imperceptible to multispectral sensors \citep{farmonov2023crop,guerri2024deep}. 

Despite its theoretical promise, HSI remains underexplored in fine-grained crop classification, primarily due to the absence of large-scale, publicly accessible crop datasets. Existing resources predominantly consist of small-scale airborne hyperspectral collections  with limited geographic diversity—far from sufficient to support deep learning's data-hungry nature. Crucially, few datasets currently combines hyperspectral data with multispectral time-series data.

To address this issue, we construct the Hierarchical Hyperspectral Crop dataset (H$^2$Crop), the first dataset integrating 30-m EnMAP hyperspectral imagery \citep{storch2023enmap,chabrillat2024enmap}  with 10-m Sentinel-2 time series in France during the 2022–2023. 
The H$^2$Crop dataset represents a significant advancement in agricultural remote sensing with four parts: (1) EnMAP hyperspectral imagery (16,344 images at 30m resolution, 218 spectral bands) capturing detailed biochemical signatures during peak growing seasons; (2) matched Sentinel-2 monthly time series (10m resolution, 12-month sequences) for phenological patterns; (3) hierarchical crop labels spanning four taxonomic levels (6, 36, 82, 101 crop types) at 10-m resolution; and (4) historical crop maps from last year's cultivation. 

H$^2$Crop addresses three critical limitations in existing datasets. First, it surpasses hyperspectral-only dataset with large-scale annotations. Second, it extends conventional LPIS-derived crop datasets (e.g., ZueriCrop) through  paired hyperspectral data. Third, it provides substantially greater spatial coverage and hierarchical granularity compared to other multi-modal datasets. The inclusion of parcel-based annotations and hyperspectral data  enhances its utility for real-world agricultural monitoring applications, establishing H$^2$Crop as an unprecedented resource for advancing fine-grained crop classification and mapping research.

In terms of methodology, we propose a dual-stream Transformer architecture that synergistically processes hyperspectral and multispectral time-series data. The hyperspectral branch employs a Spectral-Spatial Decoupled Vision Transformer, where parallel spectral and spatial transformers separately extract spectral signatures and spatial patterns. Simultaneously, the multispectral temporal branch adapts Video Swin Transformer model  to jointly extract temporal features and spatial features from 10m-resolution Sentinel-2 monthly composites, effectively capturing crop-specific temporal dynamics at field scale.

The framework achieves multi-modal integration through pixel shuffle upsampling and convolutional fusion, resolving the 30m-to-10m resolution disparity while preserving spectral-temporal features. A hierarchical classification system then processes these fused features through four cascaded prediction layers, each incorporating: (1) current spectral-temporal representations, (2) probability outputs from the preceding taxonomic level, and (3) historical crop type embeddings. This design explicitly encodes prior knowledge, maintaining phylogenetic consistency across classification hierarchies while operating efficiently through single-convolution-layer heads, resulting in simultaneous multi-granularity predictions without iterative processing.

\section{Hierarchical Hyperspectral Crop (H$^2$Crop) Dataset}
\begin{figure*}[!htb]
\centering
\includegraphics[width=\linewidth]{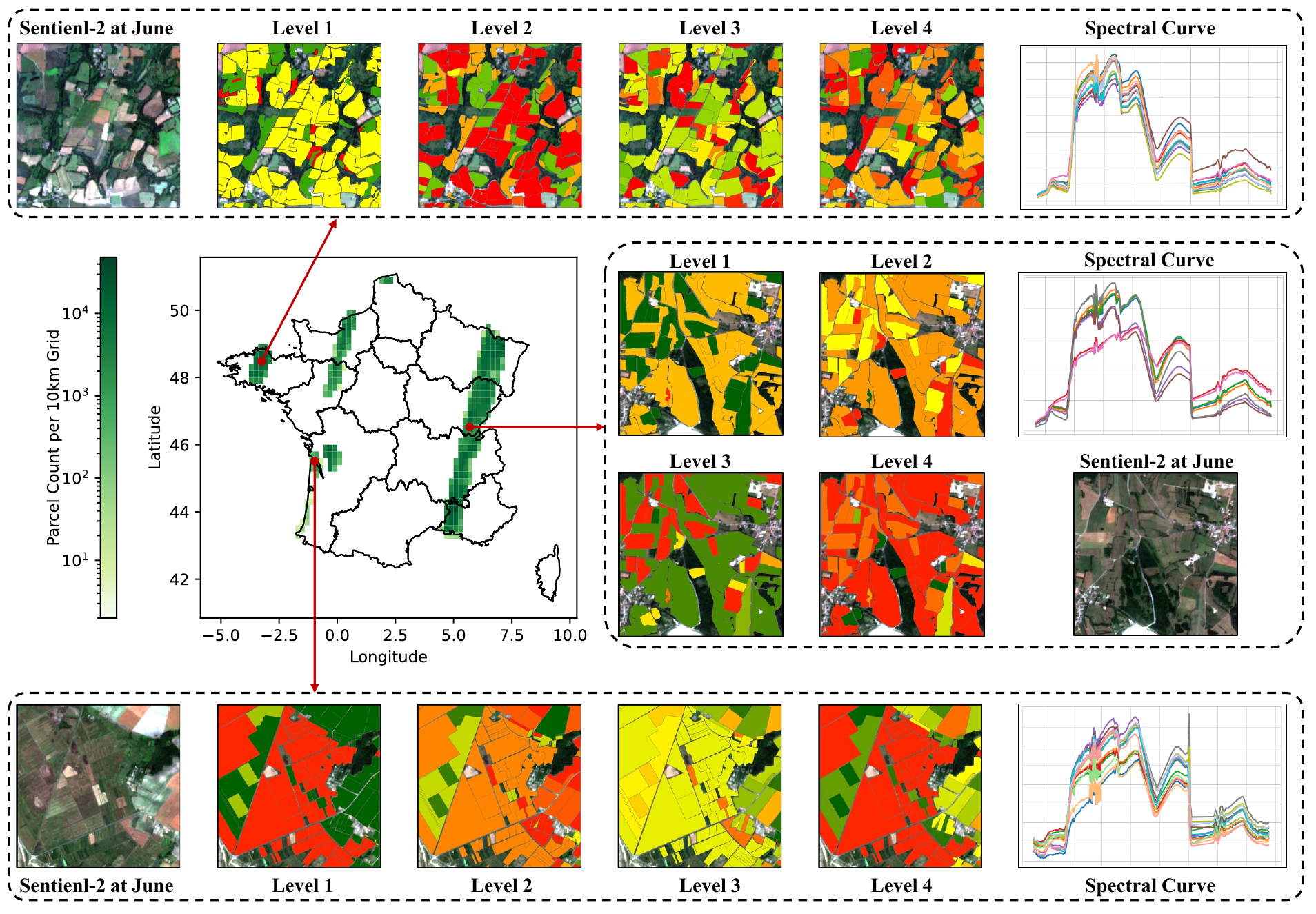}
\caption{Study area, spatial distribution and representative samples of the H$^2$Crop dataset. }
\label{fig:study_area}
\end{figure*}

The Hierarchical Hyperspectral Crop dataset (H$^2$Crop) represents the first dataset integrating time-series Sentinel-2 observations with EnMAP hyperspectral imagery for hierarchical fine-grained crop classification. It features a four-tier taxonomic system encompassing 6, 36, 82, and 101 crop types. Figure \ref{fig:study_area} illustrates the dataset’s spatial distribution across France and some representative samples, including Sentinel-2 data, exemplar crops from each hierarchical level (level 1 to level 4), and spectral curve derived from EnMAP data. 

\subsection{Materials}
\subsubsection{Satellite Data}
The multispectral data in the dataset originate from the European Space Agency (ESA)'s Sentinel-2 satellite, which provides global coverage with 13 spectral bands at spatial resolutions of 10, 20 and 60 meters. For agricultural monitoring purposes, we utilize Level-2A reflectance products, prioritizing 10m resolution bands (B2, B3, B4, B8) for spatial detail and 20m resolution bands (B5-B7, B8A, B11-B12) for enhanced spectral information in vegetation-sensitive regions.

Hyperspectral data are acquired from the Environmental Mapping and Analysis Program (EnMAP), a German hyperspectral Earth observation mission designed for environmental monitoring \citep{storch2023enmap,chabrillat2024enmap}. EnMAP delivers Level-2A atmospheric-corrected data spanning 420-1000 nm (VNIR) and 900-2450 nm (SWIR) spectral ranges, featuring 224 bands at 30m spatial resolution. To optimize vegetation analysis, we exclude bands 130-135 (1130.86-1390.48 nm) affected by atmospheric water vapor absorption and low signal-to-noise ratios, resulting in 218 validated spectral bands.

All satellite data are acquired over France during the 2022-2023 growing seasons. Sentinel-2 observations provide monthly cloud-free composites ensuring complete temporal coverage, while EnMAP acquisitions focus on critical crop growth stages in May or June.

\subsubsection{Ground Reference Data}
The ground reference data are derived from France's Land Parcel Identification System (LPIS), which serves as the foundational spatial dataset supporting the Integrated Administration and Control System (IACS) under the EU's Common Agricultural Policy (CAP). This system annually records agricultural parcels with their corresponding crop types, providing reliable ground truth for crop classification. However, the original LPIS data contains numerous categories that vary across different national implementations, requiring substantial harmonization before being suitable as deep learning labels.

The EuroCrops initiative addressed this challenge by creating a harmonized crop dataset covering multiple European countries. More importantly, it established a knowledge graph that organizes crops hierarchically according to agricultural practices through the Hierarchical Crop and Agriculture Taxonomy version 3 (HCATv3). This system encompasses approximately 400 crop types across four meaningful agricultural hierarchy levels. Each crop is assigned a unique 10-digit code where the first two digits (fixed as "33") ensure compatibility with other versions, and each subsequent two-digit pair represents progressively finer classification levels. For instance, the code 33-01-01-01-01 corresponds to ``winter common soft wheat'', whose parent level 33-01-01-01-00 represents ``common soft wheat''. This further rolls up to 33-01-01-00-00 (cereal) and ultimately to 33-01-00-00-00 (arable crops). Although initially developed for 2021 crop data, the EuroCrops framework includes mapping rules that enable consistent conversion of LPIS data from different countries and years into this standardized hierarchical classification system, allowing us to generate multi-year hierarchical crop labels.

\subsection{Dataset Construction Process}
\begin{figure*}[!htb]
\centering
\includegraphics[width=\linewidth]{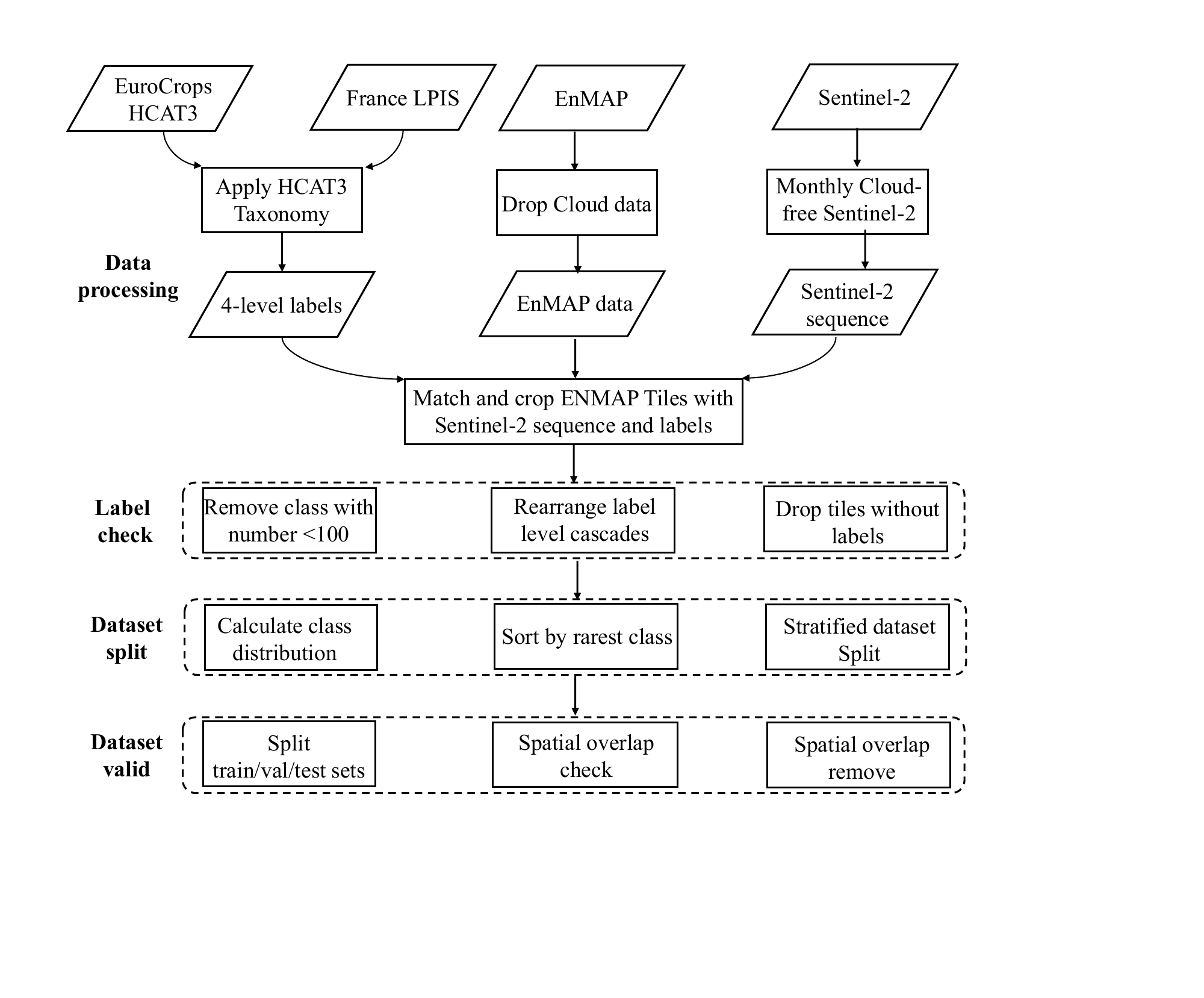}
\caption{Flowchart outlining the construction process of the H$^2$Crop dataset.}
\label{fig:flowchart}
\end{figure*}

Figure \ref{fig:flowchart} illustrates the four key steps in our dataset construction pipeline. 
First, data processing involves pre-processing three components: Sentinel-2 time series, EnMAP hyperspectral data, and LPIS ground reference. For Sentinel-2, instead of using the full 5-day revisit cycle data, we generate monthly cloud-free composites to balance dataset size and cloud contamination. The EnMAP data, acquired only during May or June, underwent rigorous cloud masking to retain only clear observations. The LPIS data are transformed into four hierarchical labels using EuroCrops' HCAT-v3 taxonomy. All data are then re-projected to a unified coordinate system and cropped to match EnMAP's coverage area, resulting in 192×192 pixels at 10m resolution for Sentinel-2 and labels, and 64×64 pixels at 30m resolution for hyperspectral data.

The second step, label check, addresses potential class imbalance caused by hyperspectral data's partial coverage and cloud filtering. We remove categories with insufficient parcels and reassign label IDs accordingly, while eliminating samples without valid crop annotations. 

For dataset splitting, we implement a frequency-aware partitioning method to handle the uneven spatial distribution of rare crops. After identifying the  least frequent crop in each image as its signature crop, we sort all samples by signature crop frequency and allocate them to training (60\%), validation (20\%), and test (20\%) sets to ensure all classes appear in each subset.

Finally, dataset validation ensured spatial independence between splits. Although EnMAP acquisitions from different dates are temporally distinct, potential spatial overlaps between training and evaluation sets should be systematically identified and removed. This rigorous four-step process yields a carefully curated, high-quality dataset for fine-grained crop classification.

\subsection{Overview of Dataset}
The H$^2$Crop dataset comprises four key components designed for fine-grained hierarchical crop classification research. Each element is described below with its technical specifications and acquisition details.

\textbf{Hyperspectral data.} The EnMAP hyperspectral data consists of cloud-free observations collected during May or June of 2022-2023 across France. Each 30m resolution image covers a 64×64 pixel area (1.92×1.92 km) and includes 218 carefully selected spectral bands after removing atmospheric absorption regions. The dataset contains 16,344 such hyperspectral scenes, providing detailed biochemical signatures but without temporal sequence information.

\textbf{Sentinel-2 time series.} For temporal phenological analysis, the Sentinel-2 time series data offers monthly cloud-free composites at 10m resolution. Each of the 16,344 locations features a 12-month sequence (192×192 pixels per timestamp), created by harmonizing native 10m and 20m bands through spatial upsampling. These time series precisely match the geographic coverage of the EnMAP observations.

\textbf{Labels.} The hierarchical labels provide four levels of crop classification granularity (6, 36, 82, and 101 crop types) at Sentinel-2's 10m resolution. This multi-tier annotation system enables both coarse and fine-grained crop type classification and mapping research.

\textbf{Historical crop types as prior knowledge.} As supplementary context, historical crop types document the previous year's cultivation patterns at identical locations. They are organized identically to the labels,with the same 4-level hierarchy and 10-meters resolution.

Table \ref{tab:datasets} compares H$^2$Crop against three existing dataset categories: (1) Hyperspectral  classification benchmarks (Indian Pines \citep{baumgardner2015220}, DFC 2018 \citep{jnh9-nz89-20}, WHU-OHS \citep{li2022whu} and QUH-Tangdaowan \citep{fu2023three}), (2) LPIS-derived Sentinel-2 time series crop classification datasets (ZueriCrop \citep{turkoglu2021crop}, AgriSen-COG \citep{selea2023agrisen}), and (3) Multi-modal hyperspectral-multispectral datasets (MDAS \citep{hu2023mdas}). Four critical aspects are compared: spectral characteristics (band count and spectral Wavelength), spatial coverage (image patch size and sample count N, ground sample distance), temporal information availability, and labels quantity and hierarchical granularity.  The parcel represents a collection of identically labeled, spatially contiguous pixels, usually in the form of polygon.

\begin{table}[!htb]
\centering
\caption{Comparative analysis of existing datasets across four key dimensions: spectral characteristics, spatial characteristics, temporal information, and label granularity.}
\label{tab:datasets}
\small
\resizebox{\linewidth}{!}{
    \begin{tabular}{ccccccccc}
    \toprule
    \multirow{2}{*}{Dataset} & \multirow{2}{*}{Platform} & \multicolumn{2}{c}{Spectral} & \multicolumn{2}{c}{Spatial}  & \multicolumn{2}{c}{Labels} & \multirow{2}{*}{Temporal}  \\
    \cmidrule(lr){3-4} \cmidrule(lr){5-6} \cmidrule(lr){7-8}
    & & Bands & Wavelength($\upmu$m) & shape([N]$\times$H$\times$W) & GSD(m)& Classes & Parcels & \\
    \midrule
    Indian Pines & Aerial: AVIRIS & 200 & 0.4-2.5 & 145$\times$145 & 20 & 16 & 40 & --\\
    DFC 2018 & Aerial: CASI-1500 & 48 & 0.38-1.05 & 4786$\times$1202 & 1.0 & 20 & 2,255 & --\\
    WHU-OHS & Satellite: Orbita & 32 & 0.4-1.0 & 7795$\times$512$\times$512 & 10 & 24 & 117,286& -- \\
    QUH-Tangdaowan & UAV: Gaiasky mini2 & 176 & 0.4-1.0 & 1740$\times$860 & 0.15 & 18 & 133 & --\\
    ZueriCrop & Satellite: Sentinel-2 & 4 & 0.49-0.83 & 28000$\times$24$\times$24 & 10 & 5-13-48 & 116,000 & \makecell{71 times \\from 2019} \\
    AgriSen-COG & Satellite: Sentinel-2 & 4 & 0.49-0.83 & 41100$\times$366$\times$366 & 10 & 102 & 6,972,485 & 2019-2020\\
    \midrule
    \multirow{3}{*}{MDAS} & Satellite: Sentinel-1 & 2 & -- & 1371$\times$888 & 10 & \multirow{3}{*}{20} & \multirow{3}{*}{32,835} \\
    & Satellite: Sentinel-2 & 12 & 0.44-2.2 & 1371$\times$888 & 10 & & & --\\
    & Aerial: HySpex  & 242 & 0.4-2.5 & 457$\times$296 & 30 & \\
    \midrule
    \multirow{2}{*}{\textbf{H$^2$Crop}} & Satellite: Sentinel-2 & 10 & 0.49-2.2 & 16344$\times$192$\times$192 & 10 & \multirow{2}{*}{6-36-82-101} & \multirow{2}{*}{1,211,418} & \multirow{2}{*}{12 per year}\\
    & Satellite: EnMAP & 218 & 0.4-2.5 & 16344$\times$64$\times$64 & 30  \\
    \bottomrule
    \end{tabular}
}
\end{table}

\begin{figure*}[!htb]
\centering
\includegraphics[width=\linewidth]{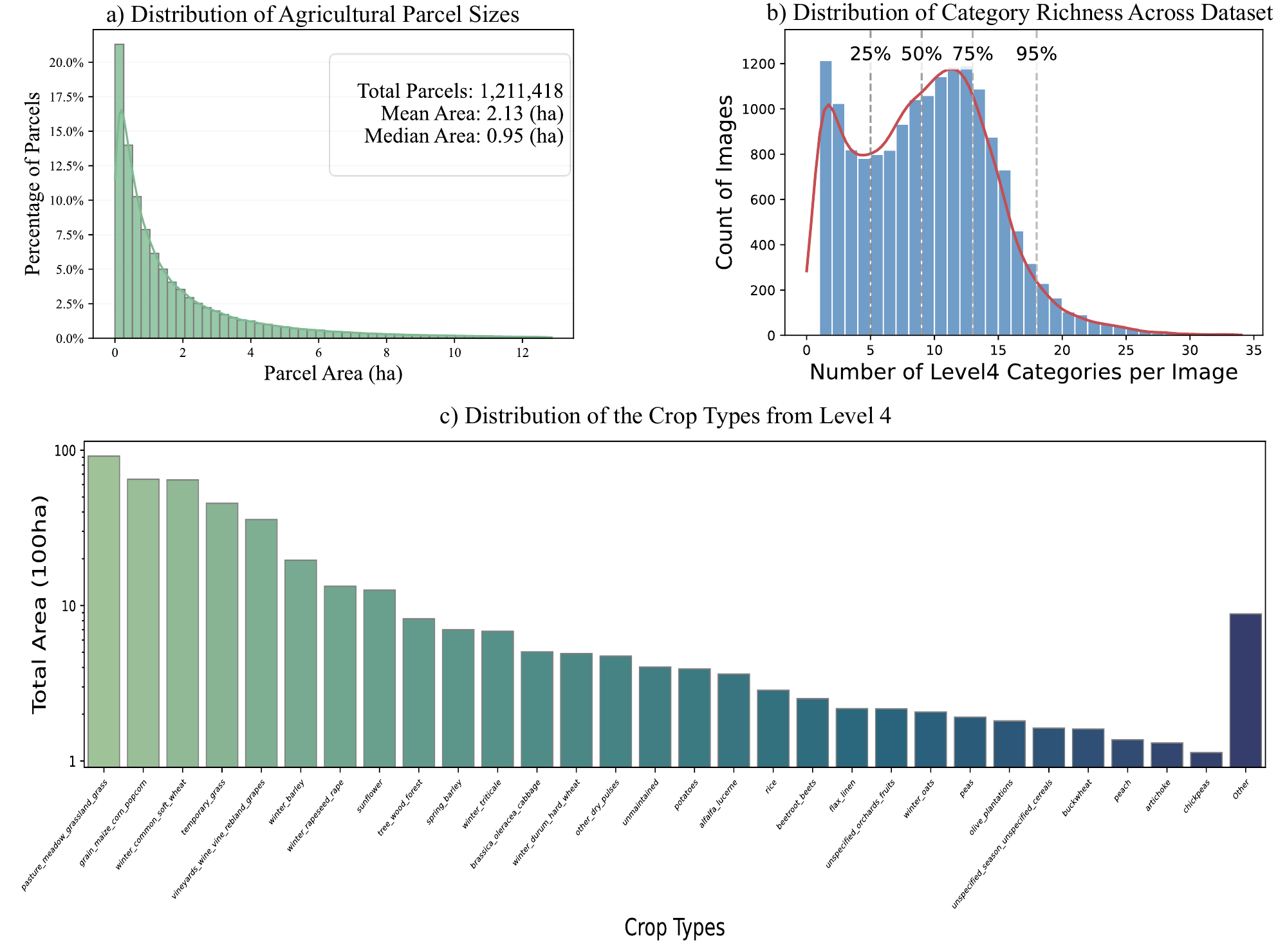}
\caption{Statistical characterization of the H$^2$Crop dataset's spatial and categorical properties.}
\label{fig:dataset_info}
\end{figure*}

The comparison reveals H$^2$Crop's unique advantages across all dimensions. Compared to conventional hyperspectral datasets, it provides an order-of-magnitude improvement in both data volume  and label refinement, making it particularly valuable for developing advanced hyperspectral classification methods. While existing crop-specific datasets like ZueriCrop offer valuable Sentinel-2 time series for phenological analysis, they fundamentally lack hyperspectral data's precise discrimination capability. The MADS dataset, though incorporating both modalities, remains limited by its small sample size  and coarse classification scheme. Collectively, H$^2$Crop establishes the first large-scale, spatio-temporally aligned, and hierarchically annotated benchmark that bridges hyperspectral data with multi-temporal data—an essential resource for advancing precision agriculture through multi-modal deep learning. 

Figure \ref{fig:dataset_info} presents key statistical insights into the H$^2$Crop dataset's composition. Figure \ref{fig:dataset_info} (a) reveals the distribution of agricultural parcel sizes across the dataset's 1,211,418 parcels, where most parcels (median area: 0.95 ha) fall below 2 hectares, though the mean area reaches 2.13 ha due to a long-tailed distribution of larger parcels. Figure \ref{fig:dataset_info} (b) quantifies the categorical complexity within individual samples, demonstrating that typical observation areas contain 7-15 distinct Level 4 crop types, reflecting exceptional biodiversity in agricultural landscapes. Figure \ref{fig:dataset_info} (c) employs logarithmic scaling to visualize the area distribution of the 19 most prevalent Level 4 crop types, with all other categories aggregated as "Others." This representation highlights the natural imbalance in crop cultivation patterns, where certain dominant species occupy disproportionately large areas compared to numerous specialty crops. Such imbalance presents significant challenges for rare crop identification, which our dataset's comprehensive hierarchical labeling and spatial coverage are specifically designed to address.

\begin{figure*}[!htb]
\centering
\includegraphics[width=0.8\linewidth]{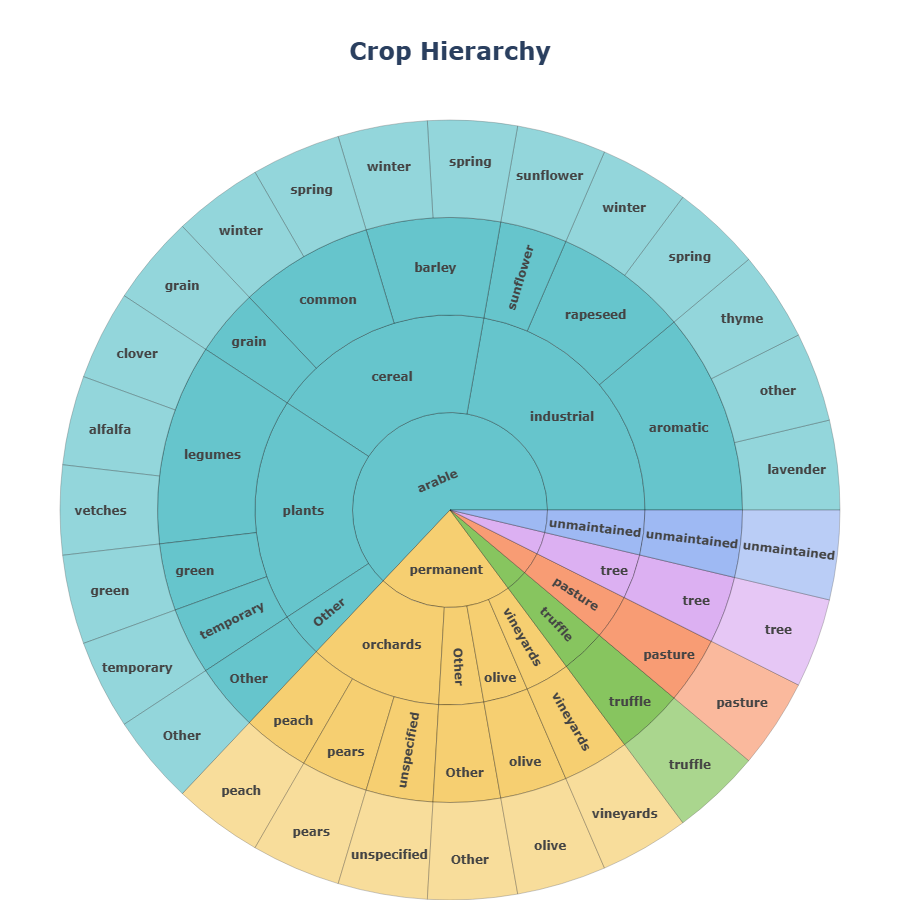}
\caption{Comprehensive four-tier hierarchical structure of agricultural taxonomy in the H$^2$Crop dataset. Only some representative crop categories are selected for clarity (complete hierarchical relationships are available in the provided code repository).}
\label{fig:cascade}
\end{figure*}

Figure \ref{fig:cascade} utilizes sunburst diagram list the four-tier classification structure of the H$^2$Crop dataset, with representative crop categories selected for clarity (complete hierarchical relationships are available in the provided code repository). The visualization reveals that arable croplands exhibit the most complex branching patterns, reflecting their extensive varietal diversity and cultivation complexity, followed by permanent crops which show moderately intricate subdivisions. This faithful representation of agricultural taxonomy provides: (1) immediate visual understanding of crop relationships across different classification levels, (2) quantitative visualization of category imbalance mirroring real-world farming practices, and (3) a reference framework for developing hierarchical classification algorithms that account for agricultural knowledge. The demonstrated taxonomy complexity particularly highlights the challenge of distinguishing closely related cultivars (e.g., wheat varieties) while validating the dataset's capacity to support fine-grained agricultural studies.

\section{Method}
\begin{figure*}[!htb]
\centering
\includegraphics[width=\linewidth]{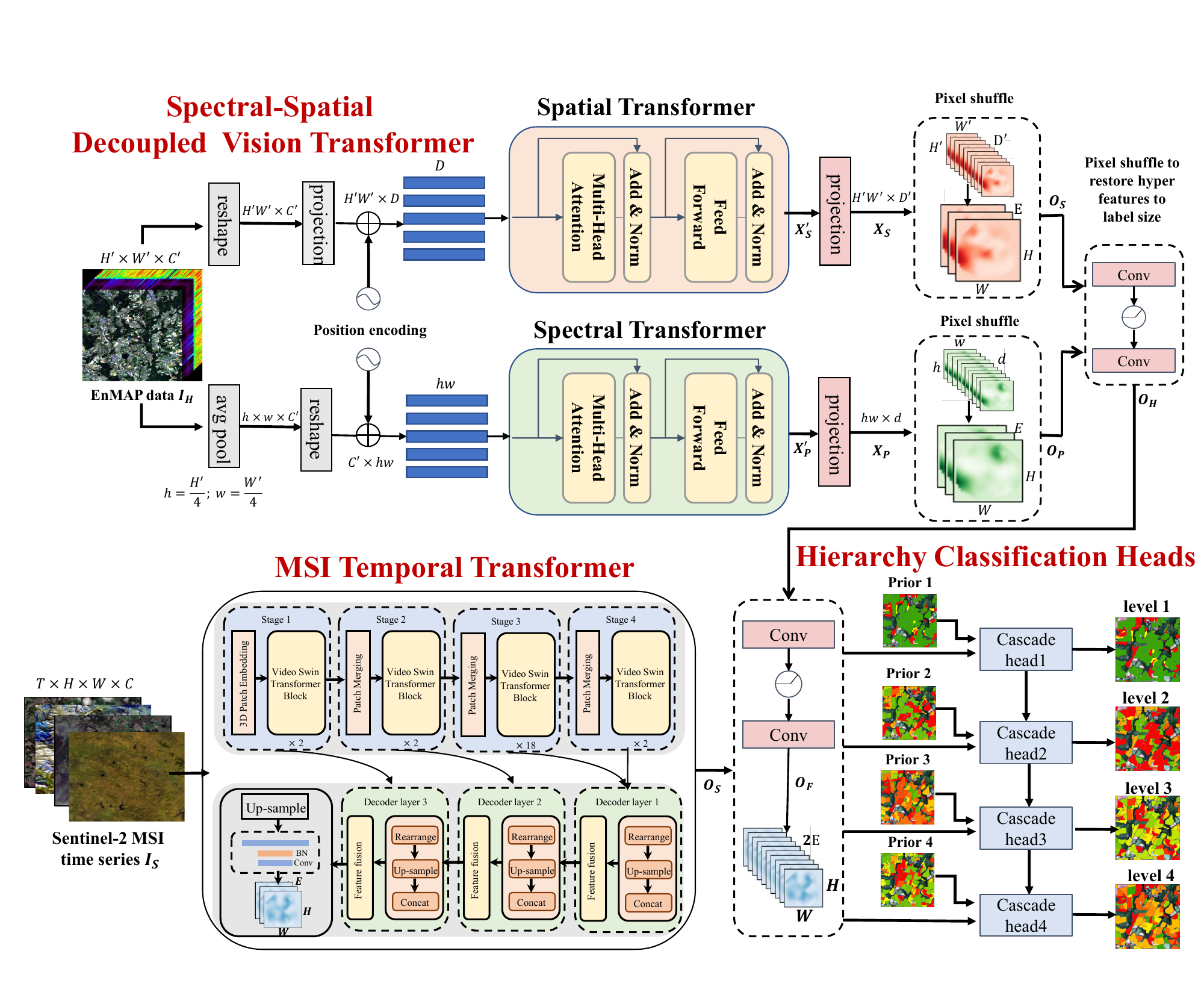}
\caption{Overall network architecture for hierarchical crop classification. Spectral and spatial transformer to extract spectral signatures and spatial patterns respectively. MSI temporal transformer to extract jointly model temporal features and spatial features from Sentinel-2 monthly composites. The  hierarchical classification heads to combine fused features, probability outputs from the preceding classification stage and historical crop type embeddings from the previous growing season for precise crop type classification.}
\label{fig:method}
\end{figure*}

We present a dual-stream Transformer architecture designed to synergistically process hyperspectral and multispectral time-series data for precise crop classification, as illustrated in Figure \ref{fig:method}. The hyperspectral branch employs a Spectral-Spatial Decoupled Vision Transformer that separately analyzes spectral signatures through spectral vision transformer and spatial patterns via spatial transformer. Simultaneously, the multispectral temporal transformer adapts Video Swin Transformer principles to jointly model temporal features and spatial features from Sentinel-2 monthly composites. These modality-specific features are aligned through pixel shuffle upsampling and integrated via convolutional layers, ensuring efficient cross-resolution fusion. The hierarchical classification heads comprises four cascaded prediction layers, each combining three information streams: (1) fused spectral-temporal features, (2) probability outputs from the preceding classification head, and (3) historical crop type embeddings from the previous growing season. This design explicitly encodes agricultural knowledge while maintaining computational efficiency through single-convolution-layer heads at each hierarchy level.

\subsection{Spectral-Spatial Decoupled Vision Transformer}

The spectral and spatial characteristics of hyperspectral data encode fundamentally distinct agricultural information: continuous spectral signatures reflect biochemical properties, while spatial patterns capture field geometry and inter-crop boundaries. To fully exploit both information types, we develop a Spectral-Spatial Decoupled Vision Transformer that separately processes spectral and spatial features, building upon existing hyperspectral classification approaches \citep{wang2022hyper,feng2024cat}.  

Given input hyperspectral data $\mathbf{I}_H \in \mathcal{R}^{H'\times W'\times C'}$ where $H'$, $W'$ and $C'$ represent height, width, and channel count, the architecture comprises two parallel transformers: spatial transformer $S_{vit}$ and spectral transformer $P_{vit}$.  

The spatial transformer $S_{vit}$ extracts spatial patterns by treating each pixel's spectral vector as a token. The input HSI data is first spatially flattened to $H'W'\times C'$, then linearly projected to dimension $D$ with added positional encodings. $N$ standard transformer layers (each is composed of multi-head self-attention, layernorm, and feed-forward networks \citep{dosovitskiy2020image}) process these tokens, producing spatial features $\mathbf{X}_S' \in \mathcal{R}^{H'W'\times D}$. These features are first reshaped to $H'\times W' \times D$ and projected through convolutional layer with increased channels, denoted as $\mathbf{X}_S\in \mathcal{R}^{H'\times W'\times D_S}$. $D_S$ is calculated as $\lfloor H/H' \rfloor \times \lfloor W/W' \rfloor \times E$, where $E$ is output dimension. Then $\mathbf{X}_S$ is upsampled in spatial sizes via pixel shuffle \citep{shi2016real,ibrahem2025pixel} - a learnable operation that rearranges depth-wise features into higher-resolution spatial outputs $\mathbf{O}_S$ (from $H'\times W'\times D_S$ to $H \times W \times E$) through periodic shuffling of channel data.

The spectral transformer $P_{vit}$ enhances spectral relationships by processing averaged patches ($h\times w \times C'$ after pooling). Flattened spectral profiles serve as tokens, with identical transformer components (multi-head attention, LayerNorm, feed-forward networks) and pixel shuffle upsampling applied. Output features $\mathbf{O}_P$ preserve spectral discriminability.  

Final feature fusion employs two convolutional layers:  
\begin{equation}
    \mathbf{O}_H=Conv(ReLu(Conv([\mathbf{O}_S;\mathbf{O}_P]))),
\end{equation}  
where $[;]$ denotes channel-wise concatenation, effectively combining spectral specificity with spatial context.  

\subsection{Multispectral Temporal Transformer}

For multispectral temporal Sentinel-2 monthly composite data $\mathbf{I}_S \in \mathcal{R}^{T\times H\times W\times C}$, both phenological dynamics from time series and spatial patterns from high-resolution imagery are essential. We adapt the Video Swin Transformer \citep{liu2022video} architecture to jointly model these spatiotemporal relationships, incorporating temporal downsampling to consolidate time-series information while maintaining spatial detail - an approach inspired by AgriFM's efficient feature aggregation for crop mapping \citep{li2025agrifmmultisourcetemporalremote}.

The architecture employs an encoder-decoder structure,  to handle high-resolution temporal data computationally efficiently while capturing multi-scale features. The encoder begins with a 3D convolutional patch embedding layer that reduces input dimensions to $\mathbf{F}_1 \in \mathcal{R}^{T_1\times H_1\times W_1 \times E_S}$, followed by four progressive transformer stages. Each stage contains $2n$ alternating transformer layers: even positions use window-based multi-head self-attention (W-MSA) for local spatiotemporal modeling within non-overlapping $M\times M\times K$ windows, while odd positions employ shifted window attention (SW-MSA) to establish cross-window connections. This dual-attention mechanism balances computational efficiency with global context awareness, where W-MSA processes isolated windows and SW-MSA recovers inter-window relationships through $\lfloor M/2 \rfloor$ shifts along spatial-temporal axes. 

Three subsequent stages apply identical processing with patch merging between stages, combining adjacent $2\times 2\times 2$ patches to halve spatial-temporal resolution while doubling feature depth ($\mathbf{F}_i \in \mathcal{R}^{T_i\times H_i \times W_i \times C_i}$ where $T_i=T_1/2^i$, $H_i=H_1/2^i$, $W_i=W_1/2^i$, $C_i=2^iE_S$). This hierarchical contraction captures phenological patterns at multiple granularities, from monthly growth stages (early stages) to seasonal trends (later stages). After each stage, the output features $\mathbf{X}_i$ with the same shape with $\mathbf{F}_i$ is served as joint spatial-temporal feature maps for further operation.

The decoder reconstructs a single semantic map through three upsampling layers, each doubling resolution via upsampling and convolutional layers. Decoder layer $i$ fuses two inputs: 1) upsampled features $\mathbf{U}_{i-1}$ from the previous layer (initialized with stage 4 encoder outputs $\mathbf{X}_4$), and 2) skip-connected features $\mathbf{F}_j$ from encoder stage $j=4-i$. Before fusion, encoder features $\mathbf{F}_j$ are reshaped to $\mathbf{F'}_j \in \mathcal{R}^{H_j\times W_j\times 2^jT_jE_S}$ to align temporal and spatial dimensions. Concatenated features undergo convolutions for efficient cross-scale integration. The final decoder output is upsampled and projected to $\mathbf{O}_S \in \mathcal{R}^ {H \times W \times E}$ for subsequent hyperspectral fusion.

\subsection{Hierarchical Classification Heads}

The hierarchical classification heads consist of four cascaded prediction layers designed to simultaneously output crop classifications at all taxonomic levels. Each head processes three information streams: (1) fused spectral-spatial-temporal features, (2) probability distributions from the preceding classification stage, and (3) one-hot encoded prior-year crop types (prior crop types). These inputs ensure agricultural consistency by combining current observations with historical cultivation patterns and progressively refined predictions.

Before entering the cascade, features from both modalities are concatenated as $\mathbf{O_F} \in \mathcal{R}^{H\times W\times 2E}$. The first classification head $G^{(1)}$ operates solely on fused features and prior crop types:
\begin{equation}
    \mathbf{P}^{(1)}=G^{(1)}([\mathbf{O_F};\mathbf{R}_1]),
\end{equation}
where $\mathbf{R}_1$ denotes level-1 prior crop encodings. Subsequent heads $G^{(k)}$ (for $k \in \{2,3,4\}$) incorporate outputs from preceding levels through:
\begin{equation}
    \mathbf{P}^{(k)}=G^{(k)}([\mathbf{P}^{(k-1)};\mathbf{O_F};\mathbf{R}_k]),
\end{equation}
enforcing hierarchical dependencies. 

The composite loss function combines cross-entropy across all levels:
\begin{equation}
    L=\sum_{k=1}^{4}\frac{1}{HW}\sum_{i,j}^{H,W}\mathbf{q}_{i,j}^{(k)}*\log \mathbf{p}_{i,j}^{(k)},
\end{equation}
with $\mathbf{q}_{i,j}^{(k)}$ representing the ground truth label at position $(i,j)$ for hierarchy level $k$. This formulation ensures balanced optimization across taxonomic granularities while maintaining pixel-wise prediction consistency.

\subsection{Implementation Details}
The spatial transformer adopts a modified ViT-base architecture, utilizing only the first six transformer layers with an embedding dimension of 768. For the spectral transformer, we apply a $4\times 4$ average pooling layer prior to processing, resulting in a reduced embedding dimension of 256, followed by six standard transformer layers. Both branches output features with dimension $E=128$ after pixel shuffle operations. The temporal transformer employs a Swin-base configuration with input dimension $E=128$. All implementations use PyTorch 2.3 and are trained on four NVIDIA L40 GPUs with a batch size of 16 ($4\times4$ patches). The learning rate follows a warmup schedule, linearly increasing from $6\times 10^{-7}$ to $6\times 10^{-5}$ over the first 1000 iterations, then decaying to $6\times 10^{-6}$ via cosine annealing. Models train for 100 epochs with metrics evaluated each epoch, retaining weights achieving the highest F1 score.

To address class imbalance, we implement comprehensive data augmentation including standard geometric transformations (random horizontal/vertical flips, rotation) combined with Cut-out \citep{devries2017improved}, which synthesizes new samples by blending patches and labels to improve minority class representation.

\subsection{Validation}

Since our main purpose  in this paper is to verify the effectiveness of HSI data and time series MSI data, we organize subsequent experiments focus on the role of hyperspectral data in different occasion with only multispectral temporal data or adding prior crop type data. Our validation strategy systematically evaluates two core hypotheses: (1) hyperspectral data provides statistically significant improvements over multispectral-temporal baselines across diverse input configurations, and (2) the proposed hierarchical architecture outperforms conventional crop classification approaches. The metrics we adopt to verify include precision (P), recall (R) and F1 scores (F1):
\begin{equation}
\begin{split}
      P = &\frac{TP}{TP + FP},\\
      R = &\frac{TP}{TP + FN},\\
      F1 =& 2 \cdot \frac{\text{P} \cdot \text{R}}{\text{P} + \text{R}},
\end{split}
\end{equation}
where $TP, FP, TN, FN$ represent the number of true positive samples, false positive samples, true negative samples and false negative samples.

First, we establish a Sentinel-2 time series (first 6 month) baseline, then incrementally introduce prior crop type maps and hyperspectral data to isolate their individual and combined contributions. This quantifies hyperspectral value under typical early-season conditions. Secondly, comparative benchmarks against CNN, CNN-LSTM, and 3D CNN architectures demonstrate our method's advantages in handling spectral-temporal complexity, while ablation studies validate the cascade heads' role in enforcing taxonomic consistency. Finally, we extend the temporal window to 8/10/12 months to assess whether hyperspectral data maintains complementary value as temporal information increases.

\section{Results}
\subsection{Hyperspectral Data Benefits in Hierarchical Crop Classification} \label{sec:hyper_benefits}
This experiment systematically evaluates the performance benefits of hyperspectral data for hierarchical crop classification under varying input conditions. We design four comparative configurations: (1) S2-only: using only Sentinel-2 time-series data as the baseline; (2) S2+Hyper: augmenting Sentinel-2 with EnMAP hyperspectral imagery; (3) S2+Prior: integrating Sentinel-2 with prior crop-type distribution maps (derived from previous-year cultivation patterns); and (4) S2+Prior+Hyper: combining all modalities. The experiment specifically focuses on early growing season scenarios (first 6 months) to simulate real-world agricultural monitoring constraints. This incremental design not only isolates the standalone utility of hyperspectral data but also reveals its synergistic effects with prior knowledge.

\begin{figure*}[!htb]
\centering
\includegraphics[width=\linewidth]{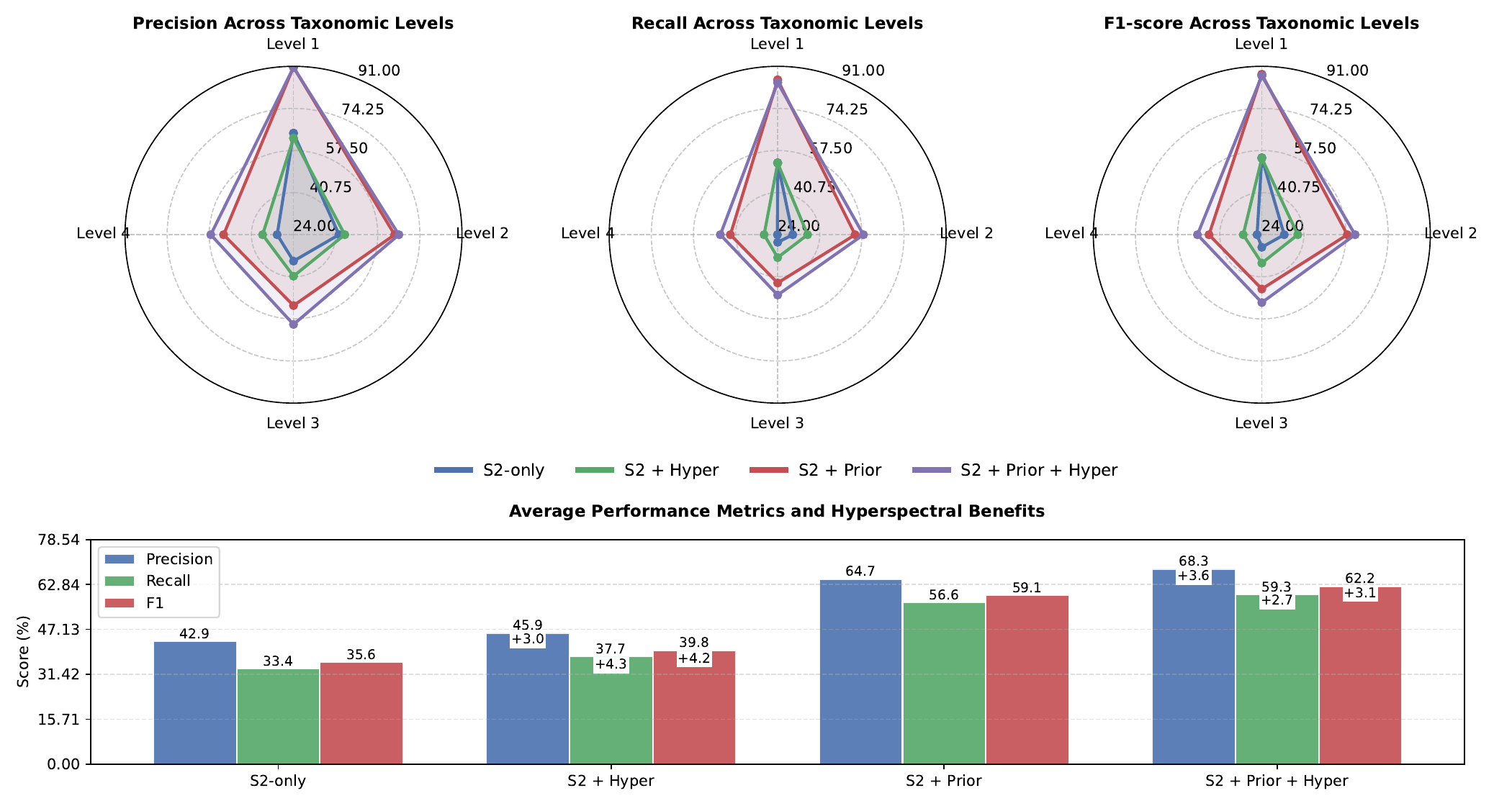}
\caption{Performance comparison of four experimental configurations. Radar charts showing precision/recall/F1 per hierarchy level. Bar plots of averaged metrics with hyperspectral contribution highlighted.}
\label{fig:hyper_benefits}
\end{figure*}

As presented in Table \ref{tab:hyper_benifits}, classification accuracy metrics, Precision (P), Recall (R), and F1 scores (F1),  are evaluated across four hierarchical levels and their averages. The "Benefit" rows quantify improvements attributable to hyperspectral inclusion. Figure \ref{fig:hyper_benefits} visualizes these results multi-dimensionally: radar charts depict precision / recall / F1 distributions for all configurations at each hierarchy level, while bar plots summarize averaged metrics with hyperspectral-induced enhancements.

\begin{table}[!htb]
\centering
\caption{Performance benefits from hyperspectral data across hierarchical levels (P: Precision, R: Recall). Benefits$\uparrow$ rows quantify absolute improvements in percentage points.}
\label{tab:hyper_benifits}
\small
\resizebox{\linewidth}{!}{
    \begin{tabular}{c|ccccccccccccccc}
    \toprule
    \multirow{2}{*}{Data} & \multicolumn{3}{c}{Average} & \multicolumn{3}{c}{Level 4 (101-class)} & \multicolumn{3}{c}{Level 3 (82-class)}  & \multicolumn{3}{c}{Level 2 (36-class)} & \multicolumn{3}{c}{Level 1 (6-class)} \\
    \cmidrule(lr){2-4} \cmidrule(lr){5-7}  \cmidrule(lr){8-10}  \cmidrule(lr){11-13}  \cmidrule(lr){14-16} 

    & P & R & F1 & P & R & F1 & P & R & F1 & P & R & F1 & P & R & F1 \\
    \midrule
    S2-only & 42.9 & 33.4 & 35.6 & 30.5 & 24.1 & 25.9 & 34.5 & 27.0 & 29.0 & 42.1 & 30.0 & 32.9 & \textbf{64.4} & \textbf{52.6} & \textbf{54.6}\\
    S2 + Hyper & \textbf{45.9} & \textbf{37.7} & \textbf{39.8} & \textbf{36.3} & \textbf{29.4} & \textbf{31.4} & \textbf{40.5} & \textbf{33.1 }& \textbf{35.3 }& \textbf{44.4} &\textbf{ 35.9} &\textbf{38.3} & 62.4 & 52.5 & 54.4 \\
    benefits $\uparrow$ &  3.0 & 4.3 &4.2 & 5.8 & 5.3 & 5.5 & 6.0 & 6.1 & 6.3 & 2.3 & 5.9 & 5.4 & -2.0 & -0.1 & -0.2\\ 
    \midrule
    S2 + Prior & 64.7 & 56.6 & 59.1 & 51.8 & 42.8 & 45.0 & 52.2 & 43.2 & 45.6 & 64.2 & 54.8 & 58.0 & 90.7 & 85.6 & 87.8\\
    S2 + Prior + Hyper & \textbf{68.3} & \textbf{59.3} & \textbf{62.2} & \textbf{56.9 }& \textbf{46.8} & \textbf{49.6} & \textbf{59.7} & \textbf{48.0 }& \textbf{51.0 }&\textbf{ 65.9} & \textbf{58.1} & \textbf{61.1} & \textbf{90.6 }& \textbf{84.5} & \textbf{87.2} \\
    benefits $\uparrow$ &  3.6 & 2.7 & 3.1 & 5.1 & 4.0 & 4.6 & 7.5 & 4.8 & 5.4 & 1.7 & 3.3 & 3.1 & -0.1 & -0.1 & -0.6\\
    \bottomrule
    \end{tabular}
}
\end{table}

Hyperspectral data delivers substantial performance gains under varying input conditions. In the baseline configuration (S2-only to S2+Hyper), incorporating hyperspectral data elevates the average F1 by 4.2\% (35.6→39.8), a statistically significant improvement. The benefits are more pronounced at finer taxonomic levels: Level 3 achieves the maximum F1 increase of 6.3\%, followed by Level 4 (+5.5\%), whereas Level 1 remains stable. This confirms that while temporal features suffice for broad-class identification, hyperspectral data provides critical discriminative power for fine-grained categorization. When prior knowledge is introduced (S2+Prior), model accuracy improves universally due to historical planting context. Even here, hyperspectral data further boosts performance by 3.1\% in average F1, with notable gains at Level 3 (+5.4\%) and Level 4 (+4.6\%), but marginal changes at Level 1. These results demonstrate that hyperspectral signatures complement prior knowledge by capturing biochemical traits inaccessible to multispectral-temporal data alone.

The consistent improvements validate our framework’s efficacy in extracting and fusing complementary information from multi-modal inputs. The Sentinel-2 processing branch effectively models spatiotemporal patterns, establishing a robust baseline, while the hyperspectral branch refines predictions through spectral fingerprinting. Our hierarchical classification head optimally leverages prior knowledge.

\begin{figure*}[!htb]
\centering
\includegraphics[width=\linewidth]{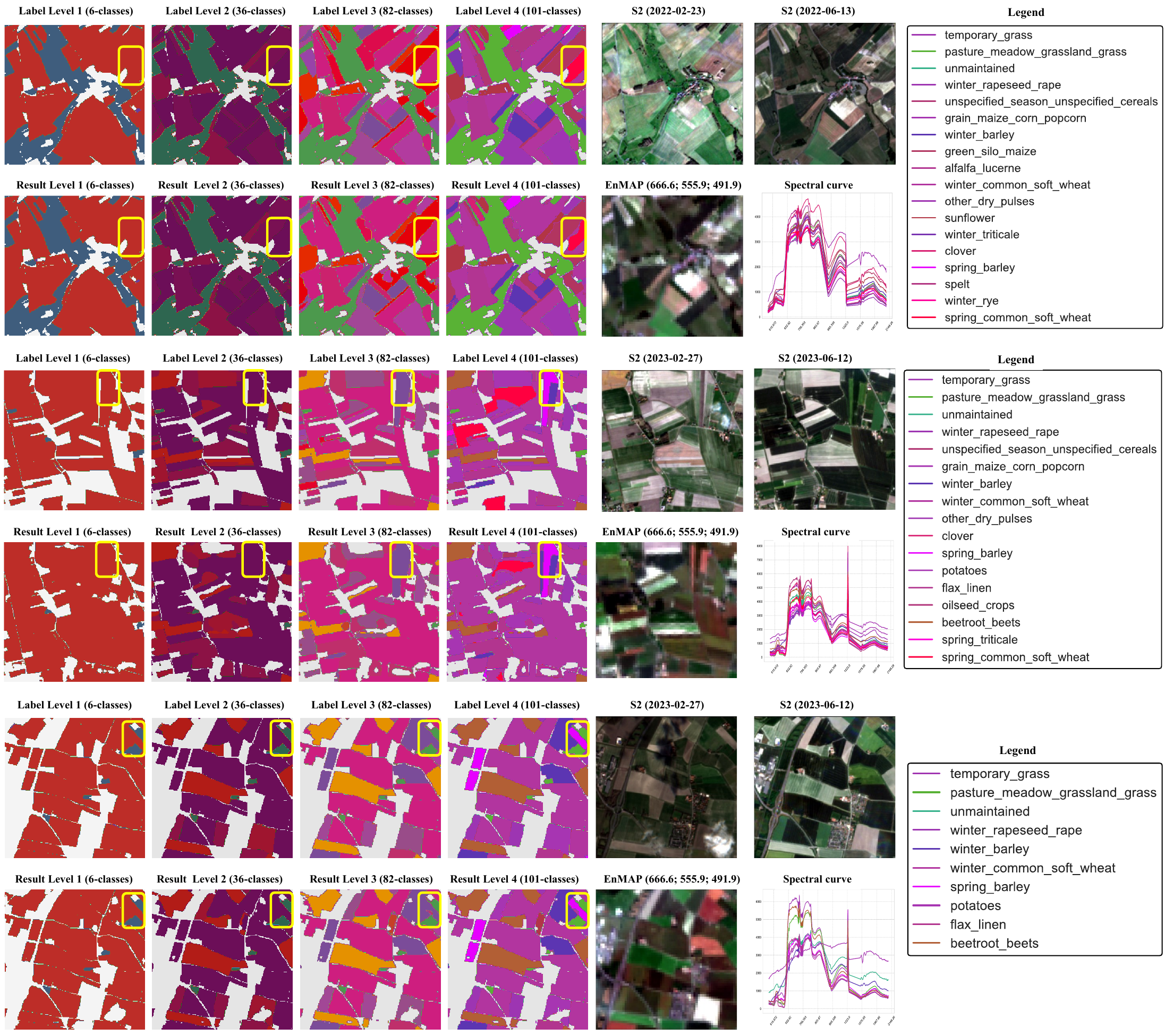}
\caption{Visualization of hierarchical classification results using S2+Prior+Hyper configuration. }
\label{fig:vis_results}
\end{figure*}

Figure \ref{fig:vis_results} illustrates three representative results under the S2+Prior+Hyper configuration. For each case, "Label Level n" displays ground truth annotations, while "Result Level n" shows model predictions. Sentinel-2 true-color composites (February and June) and EnMAP pseudo-RGB (666.6 nm, 555.9 nm, 491.9 nm) provide visual context, alongside spectral profiles of Level 4 categories.

The model accurately differentiates hierarchical classes while maintaining spatial coherence. Yellow rectangles highlight typical classification sequences from Level 1 to Level 4. In the first example, the model successfully identifies arable cropland (Level 1) → cereal (Level 2) → common soft wheat (Level 3), and precisely distinguishes spring common soft wheat from winter common soft wheat at Level 4. Such results underscore the framework’s ability to resolve taxonomically and phenologically similar crops through joint spectral-temporal hierarchical reasoning.

\subsection{Performance Comparison with Conventional Architectures}

This experiment systematically evaluates the performance of different deep learning methods for precise crop classification using full-modal data (S2+Prior+Hyper). To ensure fair comparison, all compared methods employ identical input data and training strategies but differ in feature extraction architectures: (1) UNet: uses ResNet50 \citep{he2016deep} for hyperspectral feature extraction while merging Sentinel-2 time series along the channel dimension for UNet processing; (2) 3DCNN: similarly adopts ResNet50 for hyperspectral data but processes Sentinel-2 with 3D-ResNet (replacing 2D convolutions in ResNet50 with 3D equivalents); (3) CNN-LSTM: maintains the ResNet50 hyperspectral branch and processes Sentinel-2 through CNN spatial feature extraction followed by LSTM temporal modeling. All methods share the same hierarchical classification head design. The evaluation focuses on early growing season scenarios (first 6 months) to validate practical applicability in agricultural monitoring.

\begin{figure*}[!htb]
\centering
\includegraphics[width=\linewidth]{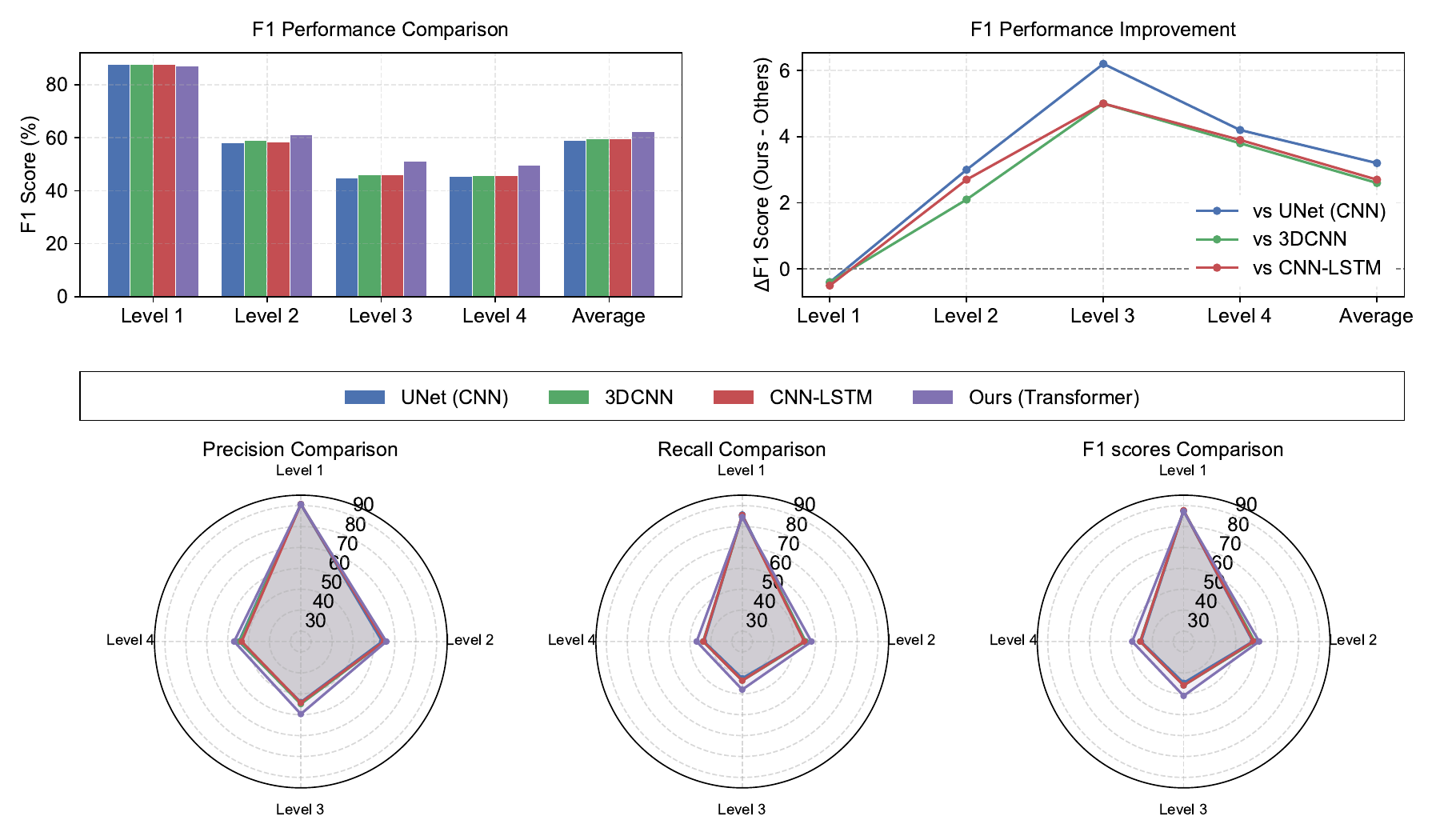}
\caption{Performance comparison of conventional deep learning  architectures: Bar charts showing F1 scores across hierarchy levels;
Line plot showing relative F1 improvement (\%) of our method over others;
Radar diagrams comparing Precision/Recall/F1 balances, where larger symmetric areas indicate more robust performance}
\label{fig:method_comparison}
\end{figure*}

Table \ref{tab:method_comparison} compares the four methods across four hierarchy levels and average performance using precision (P), recall (R), and F1 metrics. Figure \ref{fig:method_comparison} presents key findings through triple visualization: bar charts display hierarchical F1 scores, line plots quantify our method's F1 improvement over the three benchmarks, and radar plots characterize the P/R/F1 distributions across levels. Results demonstrate our method's superiority in most metrics: it achieves a 2.6 percentage-point lead in average F1 over the second-best 3DCNN (62.2 vs. 59.6), with the most pronounced advantage at fine-grained Level 4 (49.6 vs. 45.8). Notably, conventional methods show marginally better performance at coarse Level 1 (e.g., UNet's recall of 85.7 vs. our 84.5), suggesting convolutional networks' strength in global feature extraction, while our Transformer's attention mechanism excels at resolving local spectral-temporal interactions required for fine-grained discrimination. Complementing Section \ref{sec:hyper_benefits}'s findings, all methods and data combinations achieve high accuracy at Level 1 due to clear inter-class distinctions, with improvements from both architectural innovations and hyperspectral data being concentrated at finer hierarchical levels.

\begin{table}[!htb]
\centering
\caption{Performance comparison of conventional deep learning  architectures (F1/Precision/Recall metrics reported).}
\label{tab:method_comparison}
\small
\resizebox{\linewidth}{!}{
    \begin{tabular}{c|ccccccccccccccc}
    \toprule
    \multirow{2}{*}{Data} & \multicolumn{3}{c}{Average} & \multicolumn{3}{c}{Level 4 (101-class)} & \multicolumn{3}{c}{Level 3 (82-class)}  & \multicolumn{3}{c}{Level 2 (36-class)} & \multicolumn{3}{c}{Level 1 (6-class)} \\
    \cmidrule(lr){2-4} \cmidrule(lr){5-7}  \cmidrule(lr){8-10}  \cmidrule(lr){11-13}  \cmidrule(lr){14-16} 

    & P & R & F1 & P & R & F1 & P & R & F1 & P & R & F1 & P & R & F1 \\
    \midrule
    UNet (CNN) & 65.7 & 56.7 & 59.0 & 54.6 & 43.3 & 45.4 & 54.0 & 42.6 & 44.8 & 63.9 & 55.3 & 58.1 & 90.6 & \textbf{85.7} & 87.6\\
    3DCNN & 66.2 & 57.0 & 59.6 & 54.2  & 43.7 & 45.8 & 54.9  & 43.7 & 46.0 & 65.0 & 55.3 & 59.0 & \textbf{90.7} & 85.1 & 87.6\\
    CNN-LSTM & 65.8 & 56.8 & 59.5 & 53.3 & 43.6 & 45.7 & 54.3 & 43.7 & 46.0 & 64.8 & 54.7 & 58.4 & \textbf{90.7} & 85.4 & \textbf{87.7}\\
    Ours (Transformer) &  \textbf{68.3} & \textbf{59.3} & \textbf{62.2} & \textbf{56.9 }& \textbf{46.8} & \textbf{49.6} & \textbf{59.7} & \textbf{48.0 }& \textbf{51.0 }&\textbf{ 65.9} & \textbf{58.1} & \textbf{61.1} & 90.6& 84.5 & 87.2\\
    \bottomrule
    \end{tabular}
}
\end{table}

Further analysis reveals two key insights. First, our method dominates fine-grained classification (Levels 3-4), with Level 4 F1 reaching 49.6 (at least 3 points higher than alternatives), attributable to the Transformer's capacity to capture subtle spectral variations. Second, while 3DCNN slightly outperforms other benchmarks at Levels 2-3, it still trails our method by 3-5 points at finer levels, indicating 3D convolutions' limited efficacy for highly nuanced category differentiation.

Architectural analysis explains these trends: UNet's skip connections preserve multi-scale spatial features but its implicit temporal modeling (via channel expansion) causes significant F1 gaps of 4.2-5.2 points at Levels 3-4. 3DCNN's explicit 3D convolutions capture short-term spatiotemporal relationships but prove less effective than self-attention for temporal modeling. CNN-LSTM's cascaded design decouples spatial-temporal processing but suffers from LSTM's inferior long-range dependency modeling compared to Transformers. Fundamentally, these limitations confine benchmark methods to similar performance ranges (all inferior to our approach), validating our Transformer-based architecture's superiority in handling spectral-temporal agricultural data.

\subsection{Performance Comparison of Hierarchical Head and Independent Heads}

This experiment validates the effectiveness of our proposed hierarchical classification head by comparing it with independent heads, assessing the advantages of hierarchical structure in precise crop classification tasks. Independent heads refer to separately trained classifiers for each taxonomic level (e.g., Independent-L1 only predicts Level 1 categories), whereas our hierarchical head achieves end-to-end multi-level prediction through joint optimization. This design not only reduces model complexity but also enhances classification consistency through inter-level constraints.

\begin{figure*}[!htb]
\centering
\includegraphics[width=\linewidth]{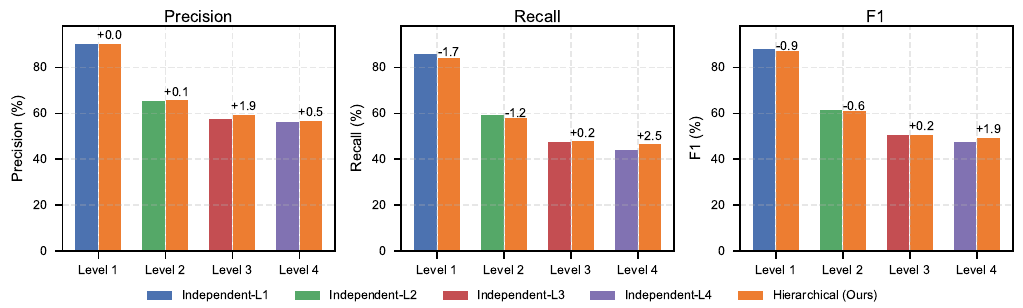}
\caption{Comparative evaluation of classification head architectures. left bars show independent head performance at corresponding levels, while right bars display our hierarchical results.}
\label{fig:hierachy}
\end{figure*}

Table \ref{tab:hier_head} compares the classification accuracy (Precision/Recall/F1) of both strategies across four hierarchical levels. Metrics for untested levels in independent heads are marked as "--". Figure \ref{fig:hierachy} further details the comparison through three sets of bar charts: for each metric group (Precision, Recall, F1), left bars show independent head performance at corresponding levels, while right bars display our hierarchical results. The experimental results demonstrate that although independent heads achieve marginal advantages at coarse levels (e.g., 0.9 percentage-point higher F1 at Level 1), our method comprehensively outperforms them at fine-grained levels (Levels 3-4). Specifically, Level 4 F1 reaches 49.6, surpassing independent heads by 1.9 percentage-points with particularly notable recall improvements (+2.5 percentage-points). This discrepancy indicates that while independent heads excel at modeling single-level features, they fail to leverage inter-level relationships, limiting fine-grained classification performance. 

\begin{table}[!htb]
\centering
\caption{Performance comparison between hierarchical and independent classification heads (Precision/Recall/F1). '--' denotes untested levels for independent heads. 'Relative Improvements $\uparrow$' indicates metric changes of our method versus independent heads.}
\label{tab:hier_head}
\small
\resizebox{\linewidth}{!}{
    \begin{tabular}{c|cccccccccccc}
        \toprule
        \multirow{2}{*}{Method} & \multicolumn{3}{c}{Level 1 (6-class)} & \multicolumn{3}{c}{Level 2 (36-class)} & \multicolumn{3}{c}{Level 3 (82-class)} & \multicolumn{3}{c}{Level 4 (101-class)} \\
        \cmidrule(lr){2-4} \cmidrule(lr){5-7}  \cmidrule(lr){8-10}  \cmidrule(lr){11-13} 
        & P & R & F1 & P & R & F1 & P & R & F1 & P & R & F1  \\
        \midrule
        Independent-L1 & \textbf{90.6} & \textbf{86.2} & \textbf{88.1} &  -- & -- & -- & -- & -- & -- & -- & -- & --  \\
        Independent-L2 & -- & -- & -- & 65.8 & \textbf{59.3 }& \textbf{61.7} & -- & -- & -- & -- & -- & --  \\
        Independent-L3 & -- & -- & -- & -- & -- & -- & 57.8 & 47.8 & 50.8& -- & -- & -- \\
        Independent-L4 & -- & -- & -- & -- & -- & -- & -- & -- & -- & 56.4 & 44.3 & 47.7\\
        \midrule
        Hierarchical (Ours)  & \textbf{90.6}& 84.5 & 87.2 &\textbf{ 65.9} & 58.1 & 61.1 & \textbf{59.7} & \textbf{48.0 }& \textbf{51.0 }  & \textbf{56.9 }& \textbf{46.8} & \textbf{49.6}\\
        \midrule
        Relative Improvements $\uparrow$ & 0.0 & -1.7 & -0.9 &  0.1 & -1.2 & -0.6 & 1.9 & 0.2 & 0.2 & 0.5 & 2.5 & 1.9\\
        \bottomrule
    \end{tabular}
}
\end{table}

The superiority of our hierarchical head manifests in two key aspects. First, joint training implicitly learns taxonomic dependencies (e.g., the logical chain "cereals → wheat → winter wheat"), reducing classification conflicts. Second, confidence scores from higher-level predictions serve as priors for subordinate levels, enhancing discrimination of ambiguous samples. Notably, the slight advantage of independent heads at coarse levels (Levels 1-2) likely stems from their exclusive focus on single tasks, but this benefit diminishes with finer classification granularity (Levels 3-4 show F1 improvements of 0.2-1.9 percentage-points). Overall, our hierarchical head achieves significant gains in fine-grained classification and logical consistency at minimal cost to coarse-level performance (0.9 F1 reduction at Level 1), better aligning with real-world agricultural decision-making requiring multi-level reasoning.

\subsection{Performance Under Different Temporal Lengths of Sentinel-2 Data}
This experiment systematically evaluates the impact of Sentinel-2 temporal sequence length on crop classification performance while verifying the consistent benefits of hyperspectral data across varying observation windows. By progressively extending the temporal coverage from June to August, October, and December, we investigate two critical questions: (1) whether more complete growing season observations improve classification accuracy, and (2) whether hyperspectral data maintains its complementary value as temporal information increases. The experimental design incorporates four temporal windows (up to June/August/October/December) and four data configurations (S2-only, +Hyper, +Prior, +Prior+Hyper), with a focus on average F1-score variations.

\begin{figure*}[!htb]
\centering
\includegraphics[width=\linewidth]{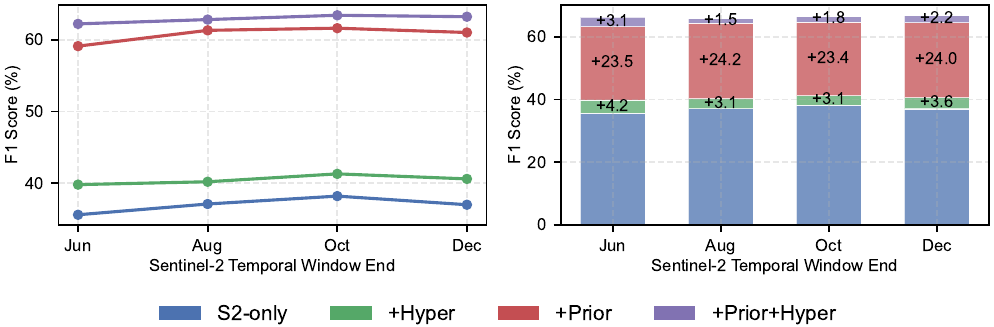}
\caption{Temporal dynamics of multi-modal data contributions: Line plot showing F1 score evolution across temporal windows for four configurations; Stacked bars showing incremental gains from prior knowledge  and hyperspectral data.}
\label{fig:temporal}
\end{figure*}

Table \ref{tab:temporal_length} presents the performance comparison across different temporal windows, with the "$\Delta_{\text{Hyper}}$" column quantifying F1-score improvements from hyperspectral data incorporation. Figure \ref{fig:temporal} employs dual visualization: a line chart illustrates the evolution of F1 scores across temporal windows for each configuration, while stacked bar charts dissect the individual contributions of Prior and Hyper at different stages. The results demonstrate that extending the observation window to October yields optimal performance (2.6 percentage-point F1 improvement for S2-only), but further extension to December may reduce accuracy due to winter-season noise. Crucially, hyperspectral data maintains significant gains across all temporal windows ($\Delta_{\text{Hyper}}$=1.5-4.2 percentage-points), with the most substantial contribution occurring in the short June window (+4.2). This robustness confirms the complementary nature of hyperspectral features to temporal data—even when complete growing season data is available, the biochemical information revealed by hyperspectral imaging remains irreplaceable.

\begin{table}[!htb]
\centering
\caption{Classification performance under different Sentinel-2 temporal windows (F1 scores reported). $\Delta_{\text{Hyper}}$ quantifies absolute F1 improvements from hyperspectral data addition.}
\label{tab:temporal_length}
\small
    \begin{tabular}{l|ccc|ccc}
    \toprule
        Temporal Window & S2-only & +Hyper &  $\Delta_{\text{Hyper}}$ & +Prior & +Prior+Hyper &  $\Delta_{\text{Hyper}}$ \\
        \midrule
        S2 (Jan-Jun) & 35.6 & 39.8 & \textbf{4.2} & 59.1 & 62.2 & \textbf{3.1}\\
        S2 (Jan-Aug) &  37.1 & 40.2 & 3.1  & 61.3 & 62.8 & 1.5\\
        S2 (Jan-Oct) & \textbf{38.2} & \textbf{41.3} & 3.1 & \textbf{61.6} & \textbf{63.4} & 1.8\\
        S2 (Jan-Dec) &  37.0 & 40.6 & 3.6 & 61.0 & 63.2 & 2.2\\
        \bottomrule
    \end{tabular}
\end{table}

Notably, hyperspectral data shows greater benefits in both short (June) and full-year (December) windows compared to intermediate periods (August-October), suggesting its dual role: compensating for insufficient temporal information in early stages while assisting in distinguishing winter crop phenotypes during later stages. This pattern underscores hyperspectral data's unique capacity to address different challenges at various phenological phases.

\section{Discussion}
\subsection{Differential Impacts of Priors and Hyperspectral Data on Changed and Unchanged Crops}

This section focuses on analyzing the differential effects of prior knowledge (prior) and hyperspectral data (hyper) in areas with changed versus unchanged crop types. By partitioning test regions into "changed" (current-year crop type differs from previous year) and "unchanged" (crop type remains consistent) categories, we thoroughly investigate the operational boundaries of these auxiliary data in practical agricultural scenarios. Here, prior derives from previous-year crop classification maps, while hyper refers to EnMAP hyperspectral imagery.

\begin{figure*}[!htb]
\centering
\includegraphics[width=\linewidth]{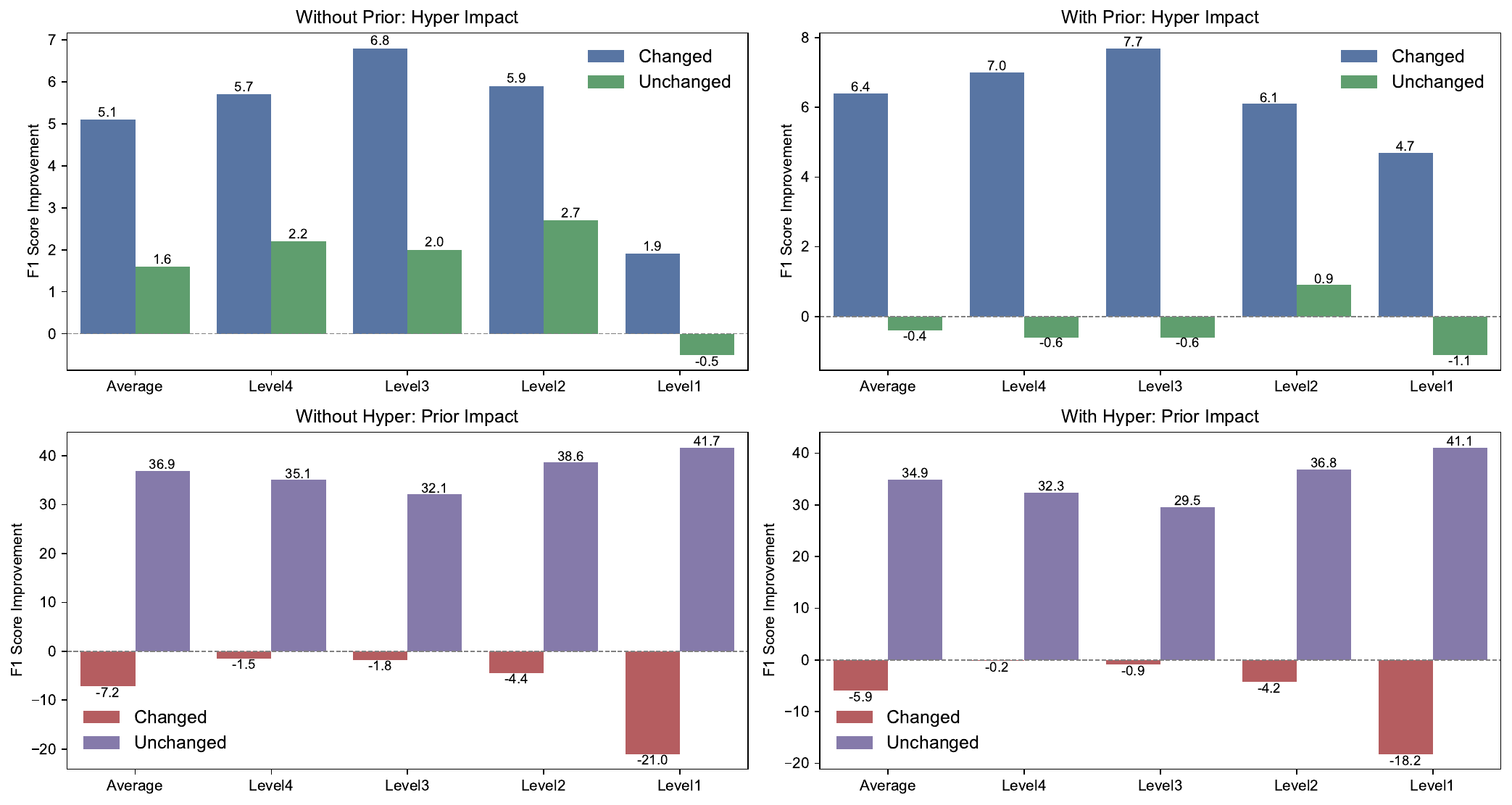}
\caption{Differential effects of hyperspectral data and prior knowledge: (Top) F1 improvements from hyperspectral data without (left) and with (right) prior knowledge; (Bottom) F1 changes from prior knowledge without (left) and with (right) hyperspectral data.}
\label{fig:changed}
\end{figure*}

Table \ref{tab:changed} details the hierarchical classification accuracy under four data configurations across changed and unchanged areas. The most striking finding reveals that in changed areas, introducing prior decreases average F1-score from 25.8 (S2-only) to 18.6 (S2+Prior), a 7.2 percentage-point degradation. Conversely, in unchanged areas, the same configuration elevates F1 from 31.1 to 68.0, a 36.9 percentage-point improvement. This extreme contrast demonstrates prior's effectiveness is highly contingent upon crop planting continuity. Figure \ref{fig:changed} further illustrates these results through matrix-style bar charts. Top row  shows F1 improvements from hyperspectral data without (left) and with (right) prior knowledge.  Bottom row shows F1 changes from prior knowledge without (left) and with (right) hyperspectral data. Key observations include: (1) hyper consistently delivers 5.1-6.4 percentage-point F1 gains in changed areas regardless of prior's presence; (2) hyper's improvements in unchanged areas are marginal (0.5-1.6 percentage-points); (3) hyper partially mitigates prior's adverse effects in changed areas (S2+Prior+Hyper outperforms S2+Prior by 6.4 percentage-points).

\begin{table}[!htb]
\centering
\caption{Performance comparison in changed and unchanged crop areas. 'Changed' denotes pixels with crop type alterations from previous year.}
\label{tab:changed}
\small
\resizebox{\linewidth}{!}{
    \begin{tabular}{c|c|ccccccccccccccc}
    \toprule
    &\multirow{2}{*}{Data} & \multicolumn{3}{c}{Average} & \multicolumn{3}{c}{Level 4 (101-class)} & \multicolumn{3}{c}{Level 3 (82-class)}  & \multicolumn{3}{c}{Level 2 (36-class)} & \multicolumn{3}{c}{Level 1 (6-class)} \\
    \cmidrule(lr){3-5} \cmidrule(lr){6-8}  \cmidrule(lr){9-11}  \cmidrule(lr){12-14}  \cmidrule(lr){15-17} 

    & & P & R & F1 & P & R & F1 & P & R & F1 & P & R & F1 & P & R & F1 \\
    \midrule
    \multirow{4}{*}{\rotatebox{90}{Changed}} & S2-only & 31.0 & 24.8 & 25.8 & 27.1 & 21.2 & 22.5 & 30.1 & 23.3 & 24.9 & 31.2 & 23.6 & 25.4 & 35.5 & 31.1 & 30.4\\
    &S2 + Hyper &  \textbf{36.1} & \textbf{29.8} & \textbf{30.9} & \textbf{33.4} & \textbf{26.5} & \textbf{28.2} & \textbf{36.9} & \textbf{29.7} & \textbf{31.7} & \textbf{35.6 }& \textbf{30.1} & \textbf{31.3} & \textbf{38.7} & \textbf{33.1} & \textbf{32.3}\\
    &S2 + Prior & 21.0 & 18.2 & 18.6 & 24.3 & 20.1 & 21.0 & 26.7 & 22.1 & 23.1 & 23.6 & 20.0 & 21.0 & 9.3 & 10.7 & 9.4\\
    &S2 + Prior + Hyper & 28.4 & 24.0 & 25.0 & 33.4 & 26.3 & 28.0 & 36.2 & 28.9 & 30.8 & 29.7 & 25.9 & 27.1 & 14.3 & 15.0 & 14.1\\
    \midrule
    \multirow{4}{*}{\rotatebox{90}{Unchanged}} & S2-only & 36.4 & 31.5 & 31.1 &  20.8 & 20.3 & 18.4 & 23.9 & 23.6 & 21.4 & 35.1 & 27.9 & 28.1 & 66.0 & 54.0 & 56.4\\
    &S2 + Hyper & 36.6 & 33.4 & 32.7 & 22.3 & 23.0 & 20.6 & 25.2 & 26.0 & 23.4 & 35.2 & 31.0 & 30.8 & 63.6 & 53.7 & 55.9\\
    &S2 + Prior &  69.9 & \textbf{68.5} & \textbf{68.0} & \textbf{55.0} & \textbf{55.1} & \textbf{53.5} & \textbf{55.1} & \textbf{55.3} & \textbf{53.5} & 70.2 & \textbf{66.7} & 66.7 & \textbf{99.4} & \textbf{97.0} & \textbf{98.1}\\
    &S2 + Prior + Hyper & \textbf{70.1} & 67.3  & 67.6 & 54.9 & 53.6 & 52.9 & \textbf{55.1} & 53.6 & 52.9 & \textbf{71.6} & 66.5 & \textbf{67.6} & 98.7 & 95.4 & 97.0\\
    \bottomrule
    \end{tabular}
}
\end{table}

Mechanistically, when crop types change, prior misleads the network into reinforcing incorrect feature associations, causing severe accuracy deterioration. Hyperspectral data counteracts this by providing objective spectral evidence—evidenced by Level 4 F1 improvements of 7.0 percentage-points (S2+Prior+Hyper vs. S2+Prior) in changed areas. In unchanged regions, priors dominates classification performance while hyperspectral data serves primarily as a fine-tuning agent (all-level variations $\le$1 percentage-point). Collectively, these results robustly demonstrate that hyper provides genuine discriminative information capable of substantially enhancing fine-grained crop classification, particularly in areas with frequent crop rotation where its contributions prove decisive. Our findings offer critical operational guidance: (1) in dynamic farming regions with frequent crop pattern changes, practitioners should prioritize hyperspectral data while avoiding prior; (2) in stable agricultural zones, prior can be selectively employed to boost performance, but must be coupled with hyperspectral data to mitigate potential negative impacts in transition areas.

\subsection{Relationship Between Class Distribution and Hierarchical Classification Accuracy}

\begin{figure*}[!htb]
\centering
\includegraphics[width=\linewidth]{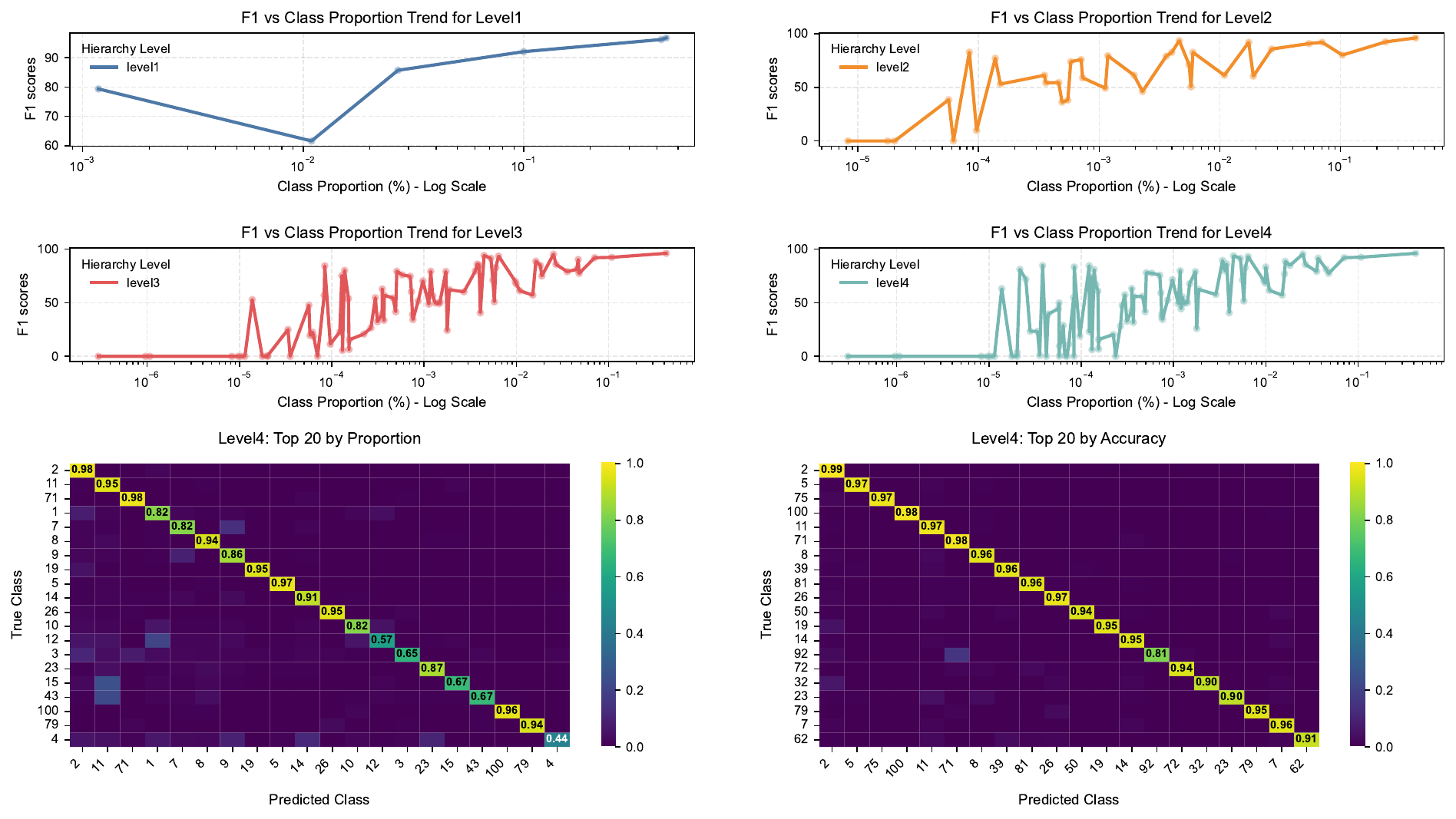}
\caption{Correlation between class distribution and classification performance. Top four line plots shows the correlation between class proportion and F1 score for Levels 1-4. The Confusion matrices show 20 most frequent classes (left) and  20 most accurately classified classes (right) in Level 4.}
\label{fig:proportion}
\end{figure*}

Figure \ref{fig:proportion} presents scatter plots with line plots showing class proportion versus F1 score for each hierarchical level. A significant positive correlation is evident across all levels - categories with larger sample sizes generally achieve higher F1 scores. However, the plots also reveal substantial performance variation among classes with similar proportions, indicating that simply increasing sample size is insufficient for fine-grained classification. This underscores the need for more sophisticated spectral-temporal feature engineering to distinguish morphologically similar crops.

The analysis includes two confusion matrices comparing the 20 most frequent classes versus the 20 most accurately classified classes in Level 4. The strong overlap between these subsets confirms our initial observation. Notably, the most abundant classes maintain relatively high accuracy.

These findings have important implications: (1) the model demonstrates superior performance for dominant crop types, aligning with operational agricultural monitoring needs; (2) the results reveal systematic biases caused by imbalanced data distribution, suggesting future work should focus on improving feature learning for rare categories; (3) from an application perspective, the method provides reliable classification for major crops, offering quantitative support for precision agriculture decisions. The persistent confusion between high-frequency similar varieties further emphasizes the need for advanced discriminative feature extraction beyond simple data quantity increases.

\subsection{Perspective and Future Directions}

The H$^2$Crop dataset and fine-grained crop classification framework proposed in this paper represent a significant breakthrough in hyperspectral agricultural remote sensing. Our dataset pioneers the synergistic integration of large-scale hyperspectral imagery (EnMAP) with multi-temporal medium-resolution Sentinel-2 data, comprising over 1 million annotated agricultural parcels across a four-tier taxonomic hierarchy. The rich spectral-temporal characteristics and meticulously curated hierarchical labels establish a new benchmark for fine-grained crop classification. Experimental results demonstrate that our refined classification architecture, which jointly models spectral features and temporal evolutionary patterns, achieves substantial improvements in crop classification accuracy - particularly proving the critical value of hyperspectral data for precision agriculture applications.

As a pioneering effort, we have validated the necessity of incorporating hyperspectral data for fine-grained crop classification and demonstrated the value of our dataset and methodology. Building upon these foundations, we identify three key directions for future advancement.

\textbf{Crop change detection paradigm}: Real-world agricultural mapping predominantly features crop type continuity (where current-year crops match previous-year plantings). Our analysis reveals fundamentally different classification behaviors in changed versus unchanged areas. Beyond serving as hierarchical classification inputs (as implemented in this paper), the included prior-year crop maps enable the derivation of dedicated change detection datasets. Transforming the task into explicit crop change detection may better align with operational needs for monitoring agricultural rotations and land use transitions \citep{chen2021adversarial,chen2024time}.

\textbf{Addressing extreme class imbalance}: Actual cropping patterns exhibit extreme class distribution disparities (up to 10,000:1 ratios). While deep learning methods inherently struggle with rare classes - potentially ignoring them entirely - we implemented conventional mitigation strategies including enhanced data augmentation. However, the pronounced imbalance in agricultural contexts demands more sophisticated solutions, possibly involving hierarchical loss reweighting or few-shot learning approaches specifically adapted for the long-tail distribution characteristics of crop species \citep{shu2019meta}.

\textbf{Advanced multi-modal interaction}: The current framework employs state-of-the-art Swin Transformers and Vision Transformers (ViT) for processing Sentinel-2 time series and hyperspectral data respectively, demonstrating proven effectiveness. However, the interaction between these modalities remains relatively simplistic. Future architectures could explore more sophisticated cross-modal attention mechanisms, hierarchical feature fusion strategies or more powerful architectures \citep{chen2024rsmamba, chen2025dynamicvis}  that better capture the complex relationships between spectral signatures and temporal growth patterns.

\section{Conclusion}
This paper pioneers an innovative paradigm for fine-grained crop classification by integrating hyperspectral data with multi-temporal satellite observations. We constructed H$^2$Crop, the first large-scale crop classification dataset combining hyperspectral and multi-temporal satellite data, representing a significant breakthrough in precision agricultural remote sensing. The H$^2$Crop dataset features precisely co-registered 30m-resolution EnMAP imagery and 10m-resolution Sentinel-2 time series, providing over one million annotated samples across a four-tier crop taxonomic hierarchy. This dataset not only fills the critical gap in hyperspectral agricultural benchmarks but also establishes an ideal benchmark for investigating spectral-temporal feature synergies through its unique multi-modal composition and hierarchical annotations.

To implement and validate this paradigm, we developed a novel multi-modal hierarchical classification architecture. Our dual-stream Transformer network extracts biochemical features from EnMAP data via  ViT branch while modeling phenological patterns in Sentinel-2 time series using a temporal Swin Transformer, ultimately integrating prior knowledge with multi-modal features through a hierarchical classification head to simultaneously generate multi-level crop classification results.

Comprehensive experiments systematically demonstrate the stable performance benefits from hyperspectral data across varying temporal spans, and crop change scenarios, with particularly notable improvements at fine-grained classification levels. These results conclusively establish the necessity of hyperspectral information for advanced crop differentiation.

The work's significance manifests across multiple dimensions. For agricultural applications, the achieved high-precision classification capability directly supports precision farming, yield estimation, and food security monitoring. Methodologically, the revealed spectral-temporal complementarity principles provide new insights for refined agricultural analytics. The open-sourced H$^2$Crop dataset will continue enabling advancements in crop identification and change detection. Particularly relevant to climate change adaptation, our method effectively captures crop feature variations under emerging cultivation patterns, offering critical technical support for climate-smart agricultural monitoring systems. Future expansions of the dataset's geographical coverage and crop diversity will further propel the transformation of agricultural remote sensing from extensive monitoring to intelligent, fine-grained observation systems.

\section*{Acknowledgment}
The research work described in this paper was conducted in the JC STEM Lab of Quantitative Remote Sensing funded by The Hong Kong Jockey Club Charities Trust. 

The computations were performed using research computing facilities offered by Information Technology Services, the University of Hong Kong.

\bibliographystyle{elsarticle-num-names}

\bibliography{refbib}

\begin{thebibliography}{70}
\expandafter\ifx\csname natexlab\endcsname\relax\def\natexlab#1{#1}\fi
\providecommand{\url}[1]{\texttt{#1}}
\providecommand{\href}[2]{#2}
\providecommand{\path}[1]{#1}
\providecommand{\DOIprefix}{doi:}
\providecommand{\ArXivprefix}{arXiv:}
\providecommand{\URLprefix}{URL: }
\providecommand{\Pubmedprefix}{pmid:}
\providecommand{\doi}[1]{\href{http://dx.doi.org/#1}{\path{#1}}}
\providecommand{\Pubmed}[1]{\href{pmid:#1}{\path{#1}}}
\providecommand{\bibinfo}[2]{#2}
\ifx\xfnm\undefined \def\xfnm[#1]{\unskip,\space#1}\fi
\bibitem[{An et~al.(2024)An, He, Zou, Yang and Zhang}]{PIS}
\bibinfo{author}{An\xfnm[ X.]}, \bibinfo{author}{He\xfnm[ W.]}, \bibinfo{author}{Zou\xfnm[ J.]}, \bibinfo{author}{Yang\xfnm[ G.]}, \bibinfo{author}{Zhang\xfnm[ H.]}.
\newblock \bibinfo{title}{Pretrain a remote sensing foundation model by promoting intra-instance similarity}.
\newblock \bibinfo{journal}{IEEE Transactions on Geoscience and Remote Sensing} \bibinfo{year}{2024};.
\bibitem[{Baumgardner et~al.(2015)Baumgardner, Biehl and Landgrebe}]{baumgardner2015220}
\bibinfo{author}{Baumgardner\xfnm[ M.]}, \bibinfo{author}{Biehl\xfnm[ L.]}, \bibinfo{author}{Landgrebe\xfnm[ D.]}.
\newblock \bibinfo{title}{220 band aviris hyperspectral image data set: June 12, 1992 indian pine test site 3}.
\newblock \bibinfo{journal}{(No Title)} \bibinfo{year}{2015};.
\bibitem[{Blickensd{\"o}rfer et~al.(2022)Blickensd{\"o}rfer, Schwieder, Pflugmacher, Nendel, Erasmi and Hostert}]{blickensdorfer2022mapping}
\bibinfo{author}{Blickensd{\"o}rfer\xfnm[ L.]}, \bibinfo{author}{Schwieder\xfnm[ M.]}, \bibinfo{author}{Pflugmacher\xfnm[ D.]}, \bibinfo{author}{Nendel\xfnm[ C.]}, \bibinfo{author}{Erasmi\xfnm[ S.]}, \bibinfo{author}{Hostert\xfnm[ P.]}.
\newblock \bibinfo{title}{Mapping of crop types and crop sequences with combined time series of sentinel-1, sentinel-2 and landsat 8 data for germany}.
\newblock \bibinfo{journal}{Remote sensing of environment} \bibinfo{year}{2022};\bibinfo{volume}{269}:\bibinfo{pages}{112831}.
\bibitem[{Breiman(2001)}]{breiman2001random}
\bibinfo{author}{Breiman\xfnm[ L.]}.
\newblock \bibinfo{title}{Random forests}.
\newblock \bibinfo{journal}{Machine learning} \bibinfo{year}{2001};\bibinfo{volume}{45}:\bibinfo{pages}{5--32}.
\bibitem[{Chabrillat et~al.(2024)Chabrillat, Foerster, Segl, Beamish, Brell, Asadzadeh, Milewski, Ward, Brosinsky, Koch et~al.}]{chabrillat2024enmap}
\bibinfo{author}{Chabrillat\xfnm[ S.]}, \bibinfo{author}{Foerster\xfnm[ S.]}, \bibinfo{author}{Segl\xfnm[ K.]}, \bibinfo{author}{Beamish\xfnm[ A.]}, \bibinfo{author}{Brell\xfnm[ M.]}, \bibinfo{author}{Asadzadeh\xfnm[ S.]}, \bibinfo{author}{Milewski\xfnm[ R.]}, \bibinfo{author}{Ward\xfnm[ K.J.]}, \bibinfo{author}{Brosinsky\xfnm[ A.]}, \bibinfo{author}{Koch\xfnm[ K.]}, et~al.
\newblock \bibinfo{title}{The enmap spaceborne imaging spectroscopy mission: Initial scientific results two years after launch}.
\newblock \bibinfo{journal}{Remote Sensing of Environment} \bibinfo{year}{2024};\bibinfo{volume}{315}:\bibinfo{pages}{114379}.
\bibitem[{Chen et~al.(2021)Chen, Li and Shi}]{chen2021adversarial}
\bibinfo{author}{Chen\xfnm[ H.]}, \bibinfo{author}{Li\xfnm[ W.]}, \bibinfo{author}{Shi\xfnm[ Z.]}.
\newblock \bibinfo{title}{Adversarial instance augmentation for building change detection in remote sensing images}.
\newblock \bibinfo{journal}{IEEE Transactions on Geoscience and Remote Sensing} \bibinfo{year}{2021};\bibinfo{volume}{60}:\bibinfo{pages}{1--16}.
\bibitem[{Chen et~al.(2024{\natexlab{a}})Chen, Chen, Liu, Li, Zou and Shi}]{chen2024rsmamba}
\bibinfo{author}{Chen\xfnm[ K.]}, \bibinfo{author}{Chen\xfnm[ B.]}, \bibinfo{author}{Liu\xfnm[ C.]}, \bibinfo{author}{Li\xfnm[ W.]}, \bibinfo{author}{Zou\xfnm[ Z.]}, \bibinfo{author}{Shi\xfnm[ Z.]}.
\newblock \bibinfo{title}{Rsmamba: Remote sensing image classification with state space model}.
\newblock \bibinfo{journal}{IEEE Geoscience and Remote Sensing Letters} \bibinfo{year}{2024}{\natexlab{a}};.
\bibitem[{Chen et~al.(2025)Chen, Liu, Chen, Li, Zou and Shi}]{chen2025dynamicvis}
\bibinfo{author}{Chen\xfnm[ K.]}, \bibinfo{author}{Liu\xfnm[ C.]}, \bibinfo{author}{Chen\xfnm[ B.]}, \bibinfo{author}{Li\xfnm[ W.]}, \bibinfo{author}{Zou\xfnm[ Z.]}, \bibinfo{author}{Shi\xfnm[ Z.]}.
\newblock \bibinfo{title}{Dynamicvis: An efficient and general visual foundation model for remote sensing image understanding}.
\newblock \bibinfo{journal}{arXiv preprint arXiv:250316426} \bibinfo{year}{2025};.
\bibitem[{Chen et~al.(2024{\natexlab{b}})Chen, Liu, Chen, Zhang, Li, Zou and Shi}]{chen2024rsprompter}
\bibinfo{author}{Chen\xfnm[ K.]}, \bibinfo{author}{Liu\xfnm[ C.]}, \bibinfo{author}{Chen\xfnm[ H.]}, \bibinfo{author}{Zhang\xfnm[ H.]}, \bibinfo{author}{Li\xfnm[ W.]}, \bibinfo{author}{Zou\xfnm[ Z.]}, \bibinfo{author}{Shi\xfnm[ Z.]}.
\newblock \bibinfo{title}{Rsprompter: Learning to prompt for remote sensing instance segmentation based on visual foundation model}.
\newblock \bibinfo{journal}{IEEE Transactions on Geoscience and Remote Sensing} \bibinfo{year}{2024}{\natexlab{b}};\bibinfo{volume}{62}:\bibinfo{pages}{1--17}.
\bibitem[{Chen et~al.(2024{\natexlab{c}})Chen, Liu, Li, Liu, Chen, Zhang, Zou and Shi}]{chen2024time}
\bibinfo{author}{Chen\xfnm[ K.]}, \bibinfo{author}{Liu\xfnm[ C.]}, \bibinfo{author}{Li\xfnm[ W.]}, \bibinfo{author}{Liu\xfnm[ Z.]}, \bibinfo{author}{Chen\xfnm[ H.]}, \bibinfo{author}{Zhang\xfnm[ H.]}, \bibinfo{author}{Zou\xfnm[ Z.]}, \bibinfo{author}{Shi\xfnm[ Z.]}.
\newblock \bibinfo{title}{Time travelling pixels: Bitemporal features integration with foundation model for remote sensing image change detection}.
\newblock In: \bibinfo{booktitle}{IGARSS 2024-2024 IEEE International Geoscience and Remote Sensing Symposium}. \bibinfo{organization}{IEEE}; \bibinfo{year}{2024}{\natexlab{c}}. p. \bibinfo{pages}{8581--8584}.
\bibitem[{Chen et~al.(2024{\natexlab{d}})Chen, Liang, Liu, Ma, Li, Sucharitakul, Luo, Chen, Fang, Zhang et~al.}]{liang2024mapping}
\bibinfo{author}{Chen\xfnm[ Y.]}, \bibinfo{author}{Liang\xfnm[ S.]}, \bibinfo{author}{Liu\xfnm[ J.]}, \bibinfo{author}{Ma\xfnm[ H.]}, \bibinfo{author}{Li\xfnm[ W.]}, \bibinfo{author}{Sucharitakul\xfnm[ P.]}, \bibinfo{author}{Luo\xfnm[ N.]}, \bibinfo{author}{Chen\xfnm[ Z.]}, \bibinfo{author}{Fang\xfnm[ H.]}, \bibinfo{author}{Zhang\xfnm[ F.]}, et~al.
\newblock \bibinfo{title}{Mapping paddy rice cropping intensity and calendar in monsoon asia at 20 m resolution between 2018 and 2021 from multi-source satellite data using a sample-free algorithm}.
\newblock \bibinfo{journal}{Available at SSRN 4948283} \bibinfo{year}{2024}{\natexlab{d}};.
\bibitem[{Delogu et~al.(2023)Delogu, Caputi, Perretta, Ripa and Boccia}]{delogu2023using}
\bibinfo{author}{Delogu\xfnm[ G.]}, \bibinfo{author}{Caputi\xfnm[ E.]}, \bibinfo{author}{Perretta\xfnm[ M.]}, \bibinfo{author}{Ripa\xfnm[ M.N.]}, \bibinfo{author}{Boccia\xfnm[ L.]}.
\newblock \bibinfo{title}{Using prisma hyperspectral data for land cover classification with artificial intelligence support}.
\newblock \bibinfo{journal}{Sustainability} \bibinfo{year}{2023};\bibinfo{volume}{15}(\bibinfo{number}{18}):\bibinfo{pages}{13786}.
\bibitem[{DeVries and Taylor(2017)}]{devries2017improved}
\bibinfo{author}{DeVries\xfnm[ T.]}, \bibinfo{author}{Taylor\xfnm[ G.W.]}.
\newblock \bibinfo{title}{Improved regularization of convolutional neural networks with cutout}.
\newblock \bibinfo{journal}{arXiv preprint arXiv:170804552} \bibinfo{year}{2017};.
\bibitem[{Dong et~al.(2020)Dong, Fu, Wang, Tian, Fu, Niu, Han, Zheng, Huang and Yuan}]{dong2020early}
\bibinfo{author}{Dong\xfnm[ J.]}, \bibinfo{author}{Fu\xfnm[ Y.]}, \bibinfo{author}{Wang\xfnm[ J.]}, \bibinfo{author}{Tian\xfnm[ H.]}, \bibinfo{author}{Fu\xfnm[ S.]}, \bibinfo{author}{Niu\xfnm[ Z.]}, \bibinfo{author}{Han\xfnm[ W.]}, \bibinfo{author}{Zheng\xfnm[ Y.]}, \bibinfo{author}{Huang\xfnm[ J.]}, \bibinfo{author}{Yuan\xfnm[ W.]}.
\newblock \bibinfo{title}{Early season mapping of winter wheat in china based on landsat and sentinel images}.
\newblock \bibinfo{journal}{Earth System Science Data Discussions} \bibinfo{year}{2020};\bibinfo{volume}{2020}:\bibinfo{pages}{1--26}.
\bibitem[{Dosovitskiy et~al.(2020)Dosovitskiy, Beyer, Kolesnikov, Weissenborn, Zhai, Unterthiner, Dehghani, Minderer, Heigold, Gelly et~al.}]{dosovitskiy2020image}
\bibinfo{author}{Dosovitskiy\xfnm[ A.]}, \bibinfo{author}{Beyer\xfnm[ L.]}, \bibinfo{author}{Kolesnikov\xfnm[ A.]}, \bibinfo{author}{Weissenborn\xfnm[ D.]}, \bibinfo{author}{Zhai\xfnm[ X.]}, \bibinfo{author}{Unterthiner\xfnm[ T.]}, \bibinfo{author}{Dehghani\xfnm[ M.]}, \bibinfo{author}{Minderer\xfnm[ M.]}, \bibinfo{author}{Heigold\xfnm[ G.]}, \bibinfo{author}{Gelly\xfnm[ S.]}, et~al.
\newblock \bibinfo{title}{An image is worth 16x16 words: Transformers for image recognition at scale}.
\newblock \bibinfo{journal}{arXiv preprint arXiv:201011929} \bibinfo{year}{2020};.
\bibitem[{Fang et~al.(2024)Fang, Liang, Chen, Ma, Li, He, Tian and Zhang}]{fang2024comprehensive}
\bibinfo{author}{Fang\xfnm[ H.]}, \bibinfo{author}{Liang\xfnm[ S.]}, \bibinfo{author}{Chen\xfnm[ Y.]}, \bibinfo{author}{Ma\xfnm[ H.]}, \bibinfo{author}{Li\xfnm[ W.]}, \bibinfo{author}{He\xfnm[ T.]}, \bibinfo{author}{Tian\xfnm[ F.]}, \bibinfo{author}{Zhang\xfnm[ F.]}.
\newblock \bibinfo{title}{A comprehensive review of rice mapping from satellite data: Algorithms, product characteristics and consistency assessment}.
\newblock \bibinfo{journal}{Science of Remote Sensing} \bibinfo{year}{2024};:\bibinfo{pages}{100172}.
\bibitem[{Fang et~al.(2025)Fang, Liang, Li, Chen, Ma, Xu, Ma, He, Tian, Zhang et~al.}]{fang5138538generating}
\bibinfo{author}{Fang\xfnm[ H.]}, \bibinfo{author}{Liang\xfnm[ S.]}, \bibinfo{author}{Li\xfnm[ W.]}, \bibinfo{author}{Chen\xfnm[ Y.]}, \bibinfo{author}{Ma\xfnm[ H.]}, \bibinfo{author}{Xu\xfnm[ J.]}, \bibinfo{author}{Ma\xfnm[ Y.]}, \bibinfo{author}{He\xfnm[ T.]}, \bibinfo{author}{Tian\xfnm[ F.]}, \bibinfo{author}{Zhang\xfnm[ F.]}, et~al.
\newblock \bibinfo{title}{Generating an annual 30m rice cover product for monsoon asia (2018-2023) using harmonized landsat and sentinel-2 data and the nasa-ibm geospatial foundation model}.
\newblock \bibinfo{journal}{Available at SSRN 5138538} \bibinfo{year}{2025};.
\bibitem[{Farmonov et~al.(2023)Farmonov, Amankulova, Szatm{\'a}ri, Sharifi, Abbasi-Moghadam, Nejad and Mucsi}]{farmonov2023crop}
\bibinfo{author}{Farmonov\xfnm[ N.]}, \bibinfo{author}{Amankulova\xfnm[ K.]}, \bibinfo{author}{Szatm{\'a}ri\xfnm[ J.]}, \bibinfo{author}{Sharifi\xfnm[ A.]}, \bibinfo{author}{Abbasi-Moghadam\xfnm[ D.]}, \bibinfo{author}{Nejad\xfnm[ S.M.M.]}, \bibinfo{author}{Mucsi\xfnm[ L.]}.
\newblock \bibinfo{title}{Crop type classification by desis hyperspectral imagery and machine learning algorithms}.
\newblock \bibinfo{journal}{IEEE Journal of selected topics in applied earth observations and remote sensing} \bibinfo{year}{2023};\bibinfo{volume}{16}:\bibinfo{pages}{1576--1588}.
\bibitem[{Fayyaz et~al.(2021)Fayyaz, Bahrami, Diba, Noroozi, Adeli, Van~Gool and Gall}]{fayyaz20213d}
\bibinfo{author}{Fayyaz\xfnm[ M.]}, \bibinfo{author}{Bahrami\xfnm[ E.]}, \bibinfo{author}{Diba\xfnm[ A.]}, \bibinfo{author}{Noroozi\xfnm[ M.]}, \bibinfo{author}{Adeli\xfnm[ E.]}, \bibinfo{author}{Van~Gool\xfnm[ L.]}, \bibinfo{author}{Gall\xfnm[ J.]}.
\newblock \bibinfo{title}{3d cnns with adaptive temporal feature resolutions}.
\newblock In: \bibinfo{booktitle}{Proceedings of the IEEE/CVF Conference on Computer Vision and Pattern Recognition}. \bibinfo{year}{2021}. p. \bibinfo{pages}{4731--4740}.
\bibitem[{Feng et~al.(2024)Feng, Wang, Zhang, Jia and Yin}]{feng2024cat}
\bibinfo{author}{Feng\xfnm[ J.]}, \bibinfo{author}{Wang\xfnm[ Q.]}, \bibinfo{author}{Zhang\xfnm[ G.]}, \bibinfo{author}{Jia\xfnm[ X.]}, \bibinfo{author}{Yin\xfnm[ J.]}.
\newblock \bibinfo{title}{Cat: Center attention transformer with stratified spatial-spectral token for hyperspectral image classification}.
\newblock \bibinfo{journal}{IEEE Transactions on Geoscience and Remote Sensing} \bibinfo{year}{2024};.
\bibitem[{Fu et~al.(2023)Fu, Sun, Zhang, Zhang, Ren, Jia and Li}]{fu2023three}
\bibinfo{author}{Fu\xfnm[ H.]}, \bibinfo{author}{Sun\xfnm[ G.]}, \bibinfo{author}{Zhang\xfnm[ L.]}, \bibinfo{author}{Zhang\xfnm[ A.]}, \bibinfo{author}{Ren\xfnm[ J.]}, \bibinfo{author}{Jia\xfnm[ X.]}, \bibinfo{author}{Li\xfnm[ F.]}.
\newblock \bibinfo{title}{Three-dimensional singular spectrum analysis for precise land cover classification from uav-borne hyperspectral benchmark datasets}.
\newblock \bibinfo{journal}{ISPRS Journal of Photogrammetry and Remote Sensing} \bibinfo{year}{2023};\bibinfo{volume}{203}:\bibinfo{pages}{115--134}.
\bibitem[{Gallo et~al.(2023)Gallo, Ranghetti, Landro, La~Grassa and Boschetti}]{gallo2023season}
\bibinfo{author}{Gallo\xfnm[ I.]}, \bibinfo{author}{Ranghetti\xfnm[ L.]}, \bibinfo{author}{Landro\xfnm[ N.]}, \bibinfo{author}{La~Grassa\xfnm[ R.]}, \bibinfo{author}{Boschetti\xfnm[ M.]}.
\newblock \bibinfo{title}{In-season and dynamic crop mapping using 3d convolution neural networks and sentinel-2 time series}.
\newblock \bibinfo{journal}{ISPRS Journal of Photogrammetry and Remote Sensing} \bibinfo{year}{2023};\bibinfo{volume}{195}:\bibinfo{pages}{335--352}.
\bibitem[{Ghojogh and Ghodsi(2023)}]{ghojogh2023recurrent}
\bibinfo{author}{Ghojogh\xfnm[ B.]}, \bibinfo{author}{Ghodsi\xfnm[ A.]}.
\newblock \bibinfo{title}{Recurrent neural networks and long short-term memory networks: Tutorial and survey}.
\newblock \bibinfo{journal}{arXiv preprint arXiv:230411461} \bibinfo{year}{2023};.
\bibitem[{Guerri et~al.(2024)Guerri, Distante, Spagnolo, Bougourzi and Taleb-Ahmed}]{guerri2024deep}
\bibinfo{author}{Guerri\xfnm[ M.F.]}, \bibinfo{author}{Distante\xfnm[ C.]}, \bibinfo{author}{Spagnolo\xfnm[ P.]}, \bibinfo{author}{Bougourzi\xfnm[ F.]}, \bibinfo{author}{Taleb-Ahmed\xfnm[ A.]}.
\newblock \bibinfo{title}{Deep learning techniques for hyperspectral image analysis in agriculture: A review}.
\newblock \bibinfo{journal}{ISPRS Open Journal of Photogrammetry and Remote Sensing} \bibinfo{year}{2024};:\bibinfo{pages}{100062}.
\bibitem[{Guo et~al.(2025)Guo, Yang, Zhang, Song, Zhang, Xu, Zhu, Ma, Wang, Bi et~al.}]{guo2025deepseek}
\bibinfo{author}{Guo\xfnm[ D.]}, \bibinfo{author}{Yang\xfnm[ D.]}, \bibinfo{author}{Zhang\xfnm[ H.]}, \bibinfo{author}{Song\xfnm[ J.]}, \bibinfo{author}{Zhang\xfnm[ R.]}, \bibinfo{author}{Xu\xfnm[ R.]}, \bibinfo{author}{Zhu\xfnm[ Q.]}, \bibinfo{author}{Ma\xfnm[ S.]}, \bibinfo{author}{Wang\xfnm[ P.]}, \bibinfo{author}{Bi\xfnm[ X.]}, et~al.
\newblock \bibinfo{title}{Deepseek-r1: Incentivizing reasoning capability in llms via reinforcement learning}.
\newblock \bibinfo{journal}{arXiv preprint arXiv:250112948} \bibinfo{year}{2025};.
\bibitem[{Han et~al.(2021)Han, Zhang, Luo, Cao, Zhang, Cheng, Zhuang, Zhang and Tao}]{han2021nesea}
\bibinfo{author}{Han\xfnm[ J.]}, \bibinfo{author}{Zhang\xfnm[ Z.]}, \bibinfo{author}{Luo\xfnm[ Y.]}, \bibinfo{author}{Cao\xfnm[ J.]}, \bibinfo{author}{Zhang\xfnm[ L.]}, \bibinfo{author}{Cheng\xfnm[ F.]}, \bibinfo{author}{Zhuang\xfnm[ H.]}, \bibinfo{author}{Zhang\xfnm[ J.]}, \bibinfo{author}{Tao\xfnm[ F.]}.
\newblock \bibinfo{title}{Nesea-rice10: high-resolution annual paddy rice maps for northeast and southeast asia from 2017 to 2019}.
\newblock \bibinfo{journal}{Earth System Science Data} \bibinfo{year}{2021};\bibinfo{volume}{13}(\bibinfo{number}{12}):\bibinfo{pages}{5969--5986}.
\bibitem[{Hara et~al.(2018)Hara, Kataoka and Satoh}]{hara2018can}
\bibinfo{author}{Hara\xfnm[ K.]}, \bibinfo{author}{Kataoka\xfnm[ H.]}, \bibinfo{author}{Satoh\xfnm[ Y.]}.
\newblock \bibinfo{title}{Can spatiotemporal 3d cnns retrace the history of 2d cnns and imagenet?}
\newblock In: \bibinfo{booktitle}{Proceedings of the IEEE conference on Computer Vision and Pattern Recognition}. \bibinfo{year}{2018}. p. \bibinfo{pages}{6546--6555}.
\bibitem[{He et~al.(2016)He, Zhang, Ren and Sun}]{he2016deep}
\bibinfo{author}{He\xfnm[ K.]}, \bibinfo{author}{Zhang\xfnm[ X.]}, \bibinfo{author}{Ren\xfnm[ S.]}, \bibinfo{author}{Sun\xfnm[ J.]}.
\newblock \bibinfo{title}{Deep residual learning for image recognition}.
\newblock In: \bibinfo{booktitle}{Proceedings of the IEEE conference on computer vision and pattern recognition}. \bibinfo{year}{2016}. p. \bibinfo{pages}{770--778}.
\bibitem[{Hearst et~al.(1998)Hearst, Dumais, Osuna, Platt and Scholkopf}]{hearst1998support}
\bibinfo{author}{Hearst\xfnm[ M.A.]}, \bibinfo{author}{Dumais\xfnm[ S.T.]}, \bibinfo{author}{Osuna\xfnm[ E.]}, \bibinfo{author}{Platt\xfnm[ J.]}, \bibinfo{author}{Scholkopf\xfnm[ B.]}.
\newblock \bibinfo{title}{Support vector machines}.
\newblock \bibinfo{journal}{IEEE Intelligent Systems and their applications} \bibinfo{year}{1998};\bibinfo{volume}{13}(\bibinfo{number}{4}):\bibinfo{pages}{18--28}.
\bibitem[{Hu et~al.(2023)Hu, Liu, Hong, Camero, Yao, Schneider, Kurz, Segl and Zhu}]{hu2023mdas}
\bibinfo{author}{Hu\xfnm[ J.]}, \bibinfo{author}{Liu\xfnm[ R.]}, \bibinfo{author}{Hong\xfnm[ D.]}, \bibinfo{author}{Camero\xfnm[ A.]}, \bibinfo{author}{Yao\xfnm[ J.]}, \bibinfo{author}{Schneider\xfnm[ M.]}, \bibinfo{author}{Kurz\xfnm[ F.]}, \bibinfo{author}{Segl\xfnm[ K.]}, \bibinfo{author}{Zhu\xfnm[ X.]}.
\newblock \bibinfo{title}{Mdas: a new multimodal benchmark dataset for remote sensing. earth system science data, 15 (1), 113--131}.
\newblock \bibinfo{year}{2023}.
\bibitem[{Ibrahem et~al.(2025)Ibrahem, Salem and Kang}]{ibrahem2025pixel}
\bibinfo{author}{Ibrahem\xfnm[ H.]}, \bibinfo{author}{Salem\xfnm[ A.]}, \bibinfo{author}{Kang\xfnm[ H.S.]}.
\newblock \bibinfo{title}{Pixel shuffling is all you need: spatially aware convmixer for dense prediction tasks}.
\newblock \bibinfo{journal}{Pattern Recognition} \bibinfo{year}{2025};\bibinfo{volume}{158}:\bibinfo{pages}{111068}.
\bibitem[{Joshi et~al.(2023)Joshi, Pradhan, Gite and Chakraborty}]{joshi2023remote}
\bibinfo{author}{Joshi\xfnm[ A.]}, \bibinfo{author}{Pradhan\xfnm[ B.]}, \bibinfo{author}{Gite\xfnm[ S.]}, \bibinfo{author}{Chakraborty\xfnm[ S.]}.
\newblock \bibinfo{title}{Remote-sensing data and deep-learning techniques in crop mapping and yield prediction: A systematic review}.
\newblock \bibinfo{journal}{Remote Sensing} \bibinfo{year}{2023};\bibinfo{volume}{15}(\bibinfo{number}{8}):\bibinfo{pages}{2014}.
\bibitem[{LeCun et~al.(2015)LeCun, Bengio and Hinton}]{lecun2015deep}
\bibinfo{author}{LeCun\xfnm[ Y.]}, \bibinfo{author}{Bengio\xfnm[ Y.]}, \bibinfo{author}{Hinton\xfnm[ G.]}.
\newblock \bibinfo{title}{Deep learning}.
\newblock \bibinfo{journal}{nature} \bibinfo{year}{2015};\bibinfo{volume}{521}(\bibinfo{number}{7553}):\bibinfo{pages}{436--444}.
\bibitem[{Li et~al.(2022)Li, Huang and Tu}]{li2022whu}
\bibinfo{author}{Li\xfnm[ J.]}, \bibinfo{author}{Huang\xfnm[ X.]}, \bibinfo{author}{Tu\xfnm[ L.]}.
\newblock \bibinfo{title}{Whu-ohs: A benchmark dataset for large-scale hersepctral image classification}.
\newblock \bibinfo{journal}{International Journal of Applied Earth Observation and Geoinformation} \bibinfo{year}{2022};\bibinfo{volume}{113}:\bibinfo{pages}{103022}.
\bibitem[{Li et~al.(2021)Li, Chen, Chen and Shi}]{li2021geographical}
\bibinfo{author}{Li\xfnm[ W.]}, \bibinfo{author}{Chen\xfnm[ K.]}, \bibinfo{author}{Chen\xfnm[ H.]}, \bibinfo{author}{Shi\xfnm[ Z.]}.
\newblock \bibinfo{title}{Geographical knowledge-driven representation learning for remote sensing images}.
\newblock \bibinfo{journal}{IEEE Transactions on Geoscience and Remote Sensing} \bibinfo{year}{2021};\bibinfo{volume}{60}:\bibinfo{pages}{1--16}.
\bibitem[{Li et~al.(2025)Li, Liang, Chen, Chen, Ma, Xu, Ma, Guan, Fang and Shi}]{li2025agrifmmultisourcetemporalremote}
\bibinfo{author}{Li\xfnm[ W.]}, \bibinfo{author}{Liang\xfnm[ S.]}, \bibinfo{author}{Chen\xfnm[ K.]}, \bibinfo{author}{Chen\xfnm[ Y.]}, \bibinfo{author}{Ma\xfnm[ H.]}, \bibinfo{author}{Xu\xfnm[ J.]}, \bibinfo{author}{Ma\xfnm[ Y.]}, \bibinfo{author}{Guan\xfnm[ S.]}, \bibinfo{author}{Fang\xfnm[ H.]}, \bibinfo{author}{Shi\xfnm[ Z.]}.
\newblock \bibinfo{title}{Agrifm: A multi-source temporal remote sensing foundation model for crop mapping}.
\newblock \bibinfo{year}{2025}.
\newblock \URLprefix \url{https://arxiv.org/abs/2505.21357}. \href{http://arxiv.org/abs/2505.21357}{\tt arXiv:2505.21357}.
\bibitem[{Li et~al.(2024)Li, Liang, Chen, Ma, Xu, Chen, Fang and Zhang}]{li5029097asiawheat}
\bibinfo{author}{Li\xfnm[ W.]}, \bibinfo{author}{Liang\xfnm[ S.]}, \bibinfo{author}{Chen\xfnm[ Y.]}, \bibinfo{author}{Ma\xfnm[ H.]}, \bibinfo{author}{Xu\xfnm[ J.]}, \bibinfo{author}{Chen\xfnm[ Z.]}, \bibinfo{author}{Fang\xfnm[ H.]}, \bibinfo{author}{Zhang\xfnm[ F.]}.
\newblock \bibinfo{title}{Asiawheat: The first asian 250-m annual fractional wheat cover time series (2001-2023) using convolutional neural networks and transformer models}.
\newblock \bibinfo{journal}{Available at SSRN 5029097} \bibinfo{year}{2024};.
\bibitem[{Liang et~al.(2024)Liang, He, Huang, Jia, Zhang, Cao, Chen, Chen, Cheng, Jiang et~al.}]{liang2024advances}
\bibinfo{author}{Liang\xfnm[ S.]}, \bibinfo{author}{He\xfnm[ T.]}, \bibinfo{author}{Huang\xfnm[ J.]}, \bibinfo{author}{Jia\xfnm[ A.]}, \bibinfo{author}{Zhang\xfnm[ Y.]}, \bibinfo{author}{Cao\xfnm[ Y.]}, \bibinfo{author}{Chen\xfnm[ X.]}, \bibinfo{author}{Chen\xfnm[ X.]}, \bibinfo{author}{Cheng\xfnm[ J.]}, \bibinfo{author}{Jiang\xfnm[ B.]}, et~al.
\newblock \bibinfo{title}{Advances in high-resolution land surface satellite products: A comprehensive review of inversion algorithms, products and challenges}.
\newblock \bibinfo{journal}{Science of Remote Sensing} \bibinfo{year}{2024};:\bibinfo{pages}{100152}.
\bibitem[{Liu et~al.(2023)Liu, Chen, Chen, Zou and Shi}]{liu2023diverse}
\bibinfo{author}{Liu\xfnm[ L.]}, \bibinfo{author}{Chen\xfnm[ B.]}, \bibinfo{author}{Chen\xfnm[ H.]}, \bibinfo{author}{Zou\xfnm[ Z.]}, \bibinfo{author}{Shi\xfnm[ Z.]}.
\newblock \bibinfo{title}{Diverse hyperspectral remote sensing image synthesis with diffusion models}.
\newblock \bibinfo{journal}{IEEE Transactions on Geoscience and Remote Sensing} \bibinfo{year}{2023};\bibinfo{volume}{61}:\bibinfo{pages}{1--16}.
\bibitem[{Liu et~al.(2021)Liu, Lin, Cao, Hu, Wei, Zhang, Lin and Guo}]{liu2021swin}
\bibinfo{author}{Liu\xfnm[ Z.]}, \bibinfo{author}{Lin\xfnm[ Y.]}, \bibinfo{author}{Cao\xfnm[ Y.]}, \bibinfo{author}{Hu\xfnm[ H.]}, \bibinfo{author}{Wei\xfnm[ Y.]}, \bibinfo{author}{Zhang\xfnm[ Z.]}, \bibinfo{author}{Lin\xfnm[ S.]}, \bibinfo{author}{Guo\xfnm[ B.]}.
\newblock \bibinfo{title}{Swin transformer: Hierarchical vision transformer using shifted windows}.
\newblock In: \bibinfo{booktitle}{Proceedings of the IEEE/CVF international conference on computer vision}. \bibinfo{year}{2021}. p. \bibinfo{pages}{10012--10022}.
\bibitem[{Liu et~al.(2022)Liu, Ning, Cao, Wei, Zhang, Lin and Hu}]{liu2022video}
\bibinfo{author}{Liu\xfnm[ Z.]}, \bibinfo{author}{Ning\xfnm[ J.]}, \bibinfo{author}{Cao\xfnm[ Y.]}, \bibinfo{author}{Wei\xfnm[ Y.]}, \bibinfo{author}{Zhang\xfnm[ Z.]}, \bibinfo{author}{Lin\xfnm[ S.]}, \bibinfo{author}{Hu\xfnm[ H.]}.
\newblock \bibinfo{title}{Video swin transformer}.
\newblock In: \bibinfo{booktitle}{Proceedings of the IEEE/CVF conference on computer vision and pattern recognition}. \bibinfo{year}{2022}. p. \bibinfo{pages}{3202--3211}.
\bibitem[{Long et~al.(2015)Long, Shelhamer and Darrell}]{long2015fully}
\bibinfo{author}{Long\xfnm[ J.]}, \bibinfo{author}{Shelhamer\xfnm[ E.]}, \bibinfo{author}{Darrell\xfnm[ T.]}.
\newblock \bibinfo{title}{Fully convolutional networks for semantic segmentation}.
\newblock In: \bibinfo{booktitle}{Proceedings of the IEEE conference on computer vision and pattern recognition}. \bibinfo{year}{2015}. p. \bibinfo{pages}{3431--3440}.
\bibitem[{Maus et~al.(2016)Maus, C{\^a}mara, Cartaxo, Sanchez, Ramos and De~Queiroz}]{maus2016time}
\bibinfo{author}{Maus\xfnm[ V.]}, \bibinfo{author}{C{\^a}mara\xfnm[ G.]}, \bibinfo{author}{Cartaxo\xfnm[ R.]}, \bibinfo{author}{Sanchez\xfnm[ A.]}, \bibinfo{author}{Ramos\xfnm[ F.M.]}, \bibinfo{author}{De~Queiroz\xfnm[ G.R.]}.
\newblock \bibinfo{title}{A time-weighted dynamic time warping method for land-use and land-cover mapping}.
\newblock \bibinfo{journal}{IEEE Journal of Selected Topics in Applied Earth Observations and Remote Sensing} \bibinfo{year}{2016};\bibinfo{volume}{9}(\bibinfo{number}{8}):\bibinfo{pages}{3729--3739}.
\bibitem[{Moharram and Sundaram(2023)}]{moharram2023land}
\bibinfo{author}{Moharram\xfnm[ M.A.]}, \bibinfo{author}{Sundaram\xfnm[ D.M.]}.
\newblock \bibinfo{title}{Land use and land cover classification with hyperspectral data: A comprehensive review of methods, challenges and future directions}.
\newblock \bibinfo{journal}{Neurocomputing} \bibinfo{year}{2023};\bibinfo{volume}{536}:\bibinfo{pages}{90--113}.
\bibitem[{Owen et~al.(2016)Owen, Milionis, Papatheodorou, Sniter, Viegas, Huth, Bortnowschi et~al.}]{owen2016land}
\bibinfo{author}{Owen\xfnm[ P.W.]}, \bibinfo{author}{Milionis\xfnm[ N.]}, \bibinfo{author}{Papatheodorou\xfnm[ I.]}, \bibinfo{author}{Sniter\xfnm[ K.]}, \bibinfo{author}{Viegas\xfnm[ H.F.]}, \bibinfo{author}{Huth\xfnm[ J.]}, \bibinfo{author}{Bortnowschi\xfnm[ R.]}, et~al.
\newblock \bibinfo{title}{The land parcel identification system: A useful tool to determine the eligibility of agricultural land—but its management could be further improved}.
\newblock \bibinfo{journal}{Special Report} \bibinfo{year}{2016};\bibinfo{volume}{25}.
\bibitem[{Peng et~al.(2023)Peng, Shen, Li, Ye, Dong, Fu and Yuan}]{peng2023twenty}
\bibinfo{author}{Peng\xfnm[ Q.]}, \bibinfo{author}{Shen\xfnm[ R.]}, \bibinfo{author}{Li\xfnm[ X.]}, \bibinfo{author}{Ye\xfnm[ T.]}, \bibinfo{author}{Dong\xfnm[ J.]}, \bibinfo{author}{Fu\xfnm[ Y.]}, \bibinfo{author}{Yuan\xfnm[ W.]}.
\newblock \bibinfo{title}{A twenty-year dataset of high-resolution maize distribution in china}.
\newblock \bibinfo{journal}{Scientific Data} \bibinfo{year}{2023};\bibinfo{volume}{10}(\bibinfo{number}{1}):\bibinfo{pages}{658}.
\bibitem[{Prasad et~al.(2020)Prasad, Le~Saux, Yokoya and Hansch}]{jnh9-nz89-20}
\bibinfo{author}{Prasad\xfnm[ S.]}, \bibinfo{author}{Le~Saux\xfnm[ B.]}, \bibinfo{author}{Yokoya\xfnm[ N.]}, \bibinfo{author}{Hansch\xfnm[ R.]}.
\newblock \bibinfo{title}{2018 ieee grss data fusion challenge – fusion of multispectral lidar and hyperspectral data}.
\newblock \bibinfo{year}{2020}.
\newblock \URLprefix \url{https://dx.doi.org/10.21227/jnh9-nz89}. \DOIprefix\doi{10.21227/jnh9-nz89}.
\bibitem[{Qiu et~al.(2022)Qiu, Hu, Chen, Tang, Yang, Zhu, Yan and Jian}]{qiu2022maps}
\bibinfo{author}{Qiu\xfnm[ B.]}, \bibinfo{author}{Hu\xfnm[ X.]}, \bibinfo{author}{Chen\xfnm[ C.]}, \bibinfo{author}{Tang\xfnm[ Z.]}, \bibinfo{author}{Yang\xfnm[ P.]}, \bibinfo{author}{Zhu\xfnm[ X.]}, \bibinfo{author}{Yan\xfnm[ C.]}, \bibinfo{author}{Jian\xfnm[ Z.]}.
\newblock \bibinfo{title}{Maps of cropping patterns in china during 2015--2021}.
\newblock \bibinfo{journal}{Scientific data} \bibinfo{year}{2022};\bibinfo{volume}{9}(\bibinfo{number}{1}):\bibinfo{pages}{479}.
\bibitem[{Ru{\ss}wurm et~al.(2023)Ru{\ss}wurm, Courty, Emonet, Lef{\`e}vre, Tuia and Tavenard}]{russwurm2023end}
\bibinfo{author}{Ru{\ss}wurm\xfnm[ M.]}, \bibinfo{author}{Courty\xfnm[ N.]}, \bibinfo{author}{Emonet\xfnm[ R.]}, \bibinfo{author}{Lef{\`e}vre\xfnm[ S.]}, \bibinfo{author}{Tuia\xfnm[ D.]}, \bibinfo{author}{Tavenard\xfnm[ R.]}.
\newblock \bibinfo{title}{End-to-end learned early classification of time series for in-season crop type mapping}.
\newblock \bibinfo{journal}{ISPRS Journal of Photogrammetry and Remote Sensing} \bibinfo{year}{2023};\bibinfo{volume}{196}:\bibinfo{pages}{445--456}.
\bibitem[{Ru{\ss}wurm et~al.(2019)Ru{\ss}wurm, Pelletier, Zollner, Lef{\`e}vre and K{\"o}rner}]{russwurm2019breizhcrops}
\bibinfo{author}{Ru{\ss}wurm\xfnm[ M.]}, \bibinfo{author}{Pelletier\xfnm[ C.]}, \bibinfo{author}{Zollner\xfnm[ M.]}, \bibinfo{author}{Lef{\`e}vre\xfnm[ S.]}, \bibinfo{author}{K{\"o}rner\xfnm[ M.]}.
\newblock \bibinfo{title}{Breizhcrops: A time series dataset for crop type mapping}.
\newblock \bibinfo{journal}{arXiv preprint arXiv:190511893} \bibinfo{year}{2019};.
\bibitem[{Schneider et~al.(2023)Schneider, Schelte, Schmitz and K{\"o}rner}]{schneider2023eurocrops}
\bibinfo{author}{Schneider\xfnm[ M.]}, \bibinfo{author}{Schelte\xfnm[ T.]}, \bibinfo{author}{Schmitz\xfnm[ F.]}, \bibinfo{author}{K{\"o}rner\xfnm[ M.]}.
\newblock \bibinfo{title}{Eurocrops: The largest harmonized open crop dataset across the european union}.
\newblock \bibinfo{journal}{Scientific Data} \bibinfo{year}{2023};\bibinfo{volume}{10}(\bibinfo{number}{1}):\bibinfo{pages}{612}.
\bibitem[{Segarra et~al.(2020)Segarra, Buchaillot, Araus and Kefauver}]{segarra2020remote}
\bibinfo{author}{Segarra\xfnm[ J.]}, \bibinfo{author}{Buchaillot\xfnm[ M.L.]}, \bibinfo{author}{Araus\xfnm[ J.L.]}, \bibinfo{author}{Kefauver\xfnm[ S.C.]}.
\newblock \bibinfo{title}{Remote sensing for precision agriculture: Sentinel-2 improved features and applications}.
\newblock \bibinfo{journal}{Agronomy} \bibinfo{year}{2020};\bibinfo{volume}{10}(\bibinfo{number}{5}):\bibinfo{pages}{641}.
\bibitem[{Selea(2023)}]{selea2023agrisen}
\bibinfo{author}{Selea\xfnm[ T.]}.
\newblock \bibinfo{title}{Agrisen-cog, a multicountry, multitemporal large-scale sentinel-2 benchmark dataset for crop mapping using deep learning}.
\newblock \bibinfo{journal}{Remote Sensing} \bibinfo{year}{2023};\bibinfo{volume}{15}(\bibinfo{number}{12}):\bibinfo{pages}{2980}.
\bibitem[{Shi et~al.(2016)Shi, Caballero, Husz{\'a}r, Totz, Aitken, Bishop, Rueckert and Wang}]{shi2016real}
\bibinfo{author}{Shi\xfnm[ W.]}, \bibinfo{author}{Caballero\xfnm[ J.]}, \bibinfo{author}{Husz{\'a}r\xfnm[ F.]}, \bibinfo{author}{Totz\xfnm[ J.]}, \bibinfo{author}{Aitken\xfnm[ A.P.]}, \bibinfo{author}{Bishop\xfnm[ R.]}, \bibinfo{author}{Rueckert\xfnm[ D.]}, \bibinfo{author}{Wang\xfnm[ Z.]}.
\newblock \bibinfo{title}{Real-time single image and video super-resolution using an efficient sub-pixel convolutional neural network}.
\newblock In: \bibinfo{booktitle}{Proceedings of the IEEE conference on computer vision and pattern recognition}. \bibinfo{year}{2016}. p. \bibinfo{pages}{1874--1883}.
\bibitem[{Shu et~al.(2019)Shu, Xie, Yi, Zhao, Zhou, Xu and Meng}]{shu2019meta}
\bibinfo{author}{Shu\xfnm[ J.]}, \bibinfo{author}{Xie\xfnm[ Q.]}, \bibinfo{author}{Yi\xfnm[ L.]}, \bibinfo{author}{Zhao\xfnm[ Q.]}, \bibinfo{author}{Zhou\xfnm[ S.]}, \bibinfo{author}{Xu\xfnm[ Z.]}, \bibinfo{author}{Meng\xfnm[ D.]}.
\newblock \bibinfo{title}{Meta-weight-net: Learning an explicit mapping for sample weighting}.
\newblock \bibinfo{journal}{Advances in neural information processing systems} \bibinfo{year}{2019};\bibinfo{volume}{32}.
\bibitem[{Storch et~al.(2023)Storch, Honold, Chabrillat, Habermeyer, Tucker, Brell, Ohndorf, Wirth, Betz, Kuchler et~al.}]{storch2023enmap}
\bibinfo{author}{Storch\xfnm[ T.]}, \bibinfo{author}{Honold\xfnm[ H.P.]}, \bibinfo{author}{Chabrillat\xfnm[ S.]}, \bibinfo{author}{Habermeyer\xfnm[ M.]}, \bibinfo{author}{Tucker\xfnm[ P.]}, \bibinfo{author}{Brell\xfnm[ M.]}, \bibinfo{author}{Ohndorf\xfnm[ A.]}, \bibinfo{author}{Wirth\xfnm[ K.]}, \bibinfo{author}{Betz\xfnm[ M.]}, \bibinfo{author}{Kuchler\xfnm[ M.]}, et~al.
\newblock \bibinfo{title}{The enmap imaging spectroscopy mission towards operations}.
\newblock \bibinfo{journal}{Remote Sensing of Environment} \bibinfo{year}{2023};\bibinfo{volume}{294}:\bibinfo{pages}{113632}.
\bibitem[{Sykas et~al.(2022)Sykas, Sdraka, Zografakis and Papoutsis}]{sykas2022sentinel}
\bibinfo{author}{Sykas\xfnm[ D.]}, \bibinfo{author}{Sdraka\xfnm[ M.]}, \bibinfo{author}{Zografakis\xfnm[ D.]}, \bibinfo{author}{Papoutsis\xfnm[ I.]}.
\newblock \bibinfo{title}{A sentinel-2 multiyear, multicountry benchmark dataset for crop classification and segmentation with deep learning}.
\newblock \bibinfo{journal}{IEEE Journal of Selected Topics in Applied Earth Observations and Remote Sensing} \bibinfo{year}{2022};\bibinfo{volume}{15}:\bibinfo{pages}{3323--3339}.
\bibitem[{Turkoglu et~al.(2021)Turkoglu, D'Aronco, Perich, Liebisch, Streit, Schindler and Wegner}]{turkoglu2021crop}
\bibinfo{author}{Turkoglu\xfnm[ M.O.]}, \bibinfo{author}{D'Aronco\xfnm[ S.]}, \bibinfo{author}{Perich\xfnm[ G.]}, \bibinfo{author}{Liebisch\xfnm[ F.]}, \bibinfo{author}{Streit\xfnm[ C.]}, \bibinfo{author}{Schindler\xfnm[ K.]}, \bibinfo{author}{Wegner\xfnm[ J.D.]}.
\newblock \bibinfo{title}{Crop mapping from image time series: Deep learning with multi-scale label hierarchies}.
\newblock \bibinfo{journal}{Remote Sensing of Environment} \bibinfo{year}{2021};\bibinfo{volume}{264}:\bibinfo{pages}{112603}.
\bibitem[{Van~Tricht et~al.(2023)Van~Tricht, Degerickx, Gilliams, Zanaga, Battude, Grosu, Brombacher, Lesiv, Bayas, Karanam et~al.}]{van2023worldcereal}
\bibinfo{author}{Van~Tricht\xfnm[ K.]}, \bibinfo{author}{Degerickx\xfnm[ J.]}, \bibinfo{author}{Gilliams\xfnm[ S.]}, \bibinfo{author}{Zanaga\xfnm[ D.]}, \bibinfo{author}{Battude\xfnm[ M.]}, \bibinfo{author}{Grosu\xfnm[ A.]}, \bibinfo{author}{Brombacher\xfnm[ J.]}, \bibinfo{author}{Lesiv\xfnm[ M.]}, \bibinfo{author}{Bayas\xfnm[ J.C.L.]}, \bibinfo{author}{Karanam\xfnm[ S.]}, et~al.
\newblock \bibinfo{title}{Worldcereal: a dynamic open-source system for global-scale, seasonal, and reproducible crop and irrigation mapping}.
\newblock \bibinfo{journal}{Earth System Science Data} \bibinfo{year}{2023};\bibinfo{volume}{15}(\bibinfo{number}{12}):\bibinfo{pages}{5491--5515}.
\bibitem[{Vaswani et~al.(2017)Vaswani, Shazeer, Parmar, Uszkoreit, Jones, Gomez, Kaiser and Polosukhin}]{vaswani2017attention}
\bibinfo{author}{Vaswani\xfnm[ A.]}, \bibinfo{author}{Shazeer\xfnm[ N.]}, \bibinfo{author}{Parmar\xfnm[ N.]}, \bibinfo{author}{Uszkoreit\xfnm[ J.]}, \bibinfo{author}{Jones\xfnm[ L.]}, \bibinfo{author}{Gomez\xfnm[ A.N.]}, \bibinfo{author}{Kaiser\xfnm[ {\L}.]}, \bibinfo{author}{Polosukhin\xfnm[ I.]}.
\newblock \bibinfo{title}{Attention is all you need}.
\newblock \bibinfo{journal}{Advances in neural information processing systems} \bibinfo{year}{2017};\bibinfo{volume}{30}.
\bibitem[{Victor et~al.(2024)Victor, Maddikunta, Mary, Murugan, Chengoden, Gadekallu, Rakesh, Zhu and Paek}]{victor2024remote}
\bibinfo{author}{Victor\xfnm[ N.]}, \bibinfo{author}{Maddikunta\xfnm[ P.K.R.]}, \bibinfo{author}{Mary\xfnm[ D.R.K.]}, \bibinfo{author}{Murugan\xfnm[ R.]}, \bibinfo{author}{Chengoden\xfnm[ R.]}, \bibinfo{author}{Gadekallu\xfnm[ T.R.]}, \bibinfo{author}{Rakesh\xfnm[ N.]}, \bibinfo{author}{Zhu\xfnm[ Y.]}, \bibinfo{author}{Paek\xfnm[ J.]}.
\newblock \bibinfo{title}{Remote sensing for agriculture in the era of industry 5.0--a survey}.
\newblock \bibinfo{journal}{IEEE Journal of Selected Topics in Applied Earth Observations and Remote Sensing} \bibinfo{year}{2024};.
\bibitem[{Wang et~al.(2022)Wang, Liu, Zhang, Shen, Wang and Li}]{wang2022hyper}
\bibinfo{author}{Wang\xfnm[ W.]}, \bibinfo{author}{Liu\xfnm[ L.]}, \bibinfo{author}{Zhang\xfnm[ T.]}, \bibinfo{author}{Shen\xfnm[ J.]}, \bibinfo{author}{Wang\xfnm[ J.]}, \bibinfo{author}{Li\xfnm[ J.]}.
\newblock \bibinfo{title}{Hyper-es2t: efficient spatial--spectral transformer for the classification of hyperspectral remote sensing images}.
\newblock \bibinfo{journal}{International Journal of Applied Earth Observation and Geoinformation} \bibinfo{year}{2022};\bibinfo{volume}{113}:\bibinfo{pages}{103005}.
\bibitem[{Weikmann et~al.(2021)Weikmann, Paris and Bruzzone}]{weikmann2021timesen2crop}
\bibinfo{author}{Weikmann\xfnm[ G.]}, \bibinfo{author}{Paris\xfnm[ C.]}, \bibinfo{author}{Bruzzone\xfnm[ L.]}.
\newblock \bibinfo{title}{Timesen2crop: A million labeled samples dataset of sentinel 2 image time series for crop-type classification}.
\newblock \bibinfo{journal}{IEEE Journal of Selected Topics in Applied Earth Observations and Remote Sensing} \bibinfo{year}{2021};\bibinfo{volume}{14}:\bibinfo{pages}{4699--4708}.
\bibitem[{Weiss et~al.(2020)Weiss, Jacob and Duveiller}]{weiss2020remote}
\bibinfo{author}{Weiss\xfnm[ M.]}, \bibinfo{author}{Jacob\xfnm[ F.]}, \bibinfo{author}{Duveiller\xfnm[ G.]}.
\newblock \bibinfo{title}{Remote sensing for agricultural applications: A meta-review}.
\newblock \bibinfo{journal}{Remote sensing of environment} \bibinfo{year}{2020};\bibinfo{volume}{236}:\bibinfo{pages}{111402}.
\bibitem[{You et~al.(2021)You, Dong, Huang, Du, Zhang, He, Yang, Di and Xiao}]{you202110}
\bibinfo{author}{You\xfnm[ N.]}, \bibinfo{author}{Dong\xfnm[ J.]}, \bibinfo{author}{Huang\xfnm[ J.]}, \bibinfo{author}{Du\xfnm[ G.]}, \bibinfo{author}{Zhang\xfnm[ G.]}, \bibinfo{author}{He\xfnm[ Y.]}, \bibinfo{author}{Yang\xfnm[ T.]}, \bibinfo{author}{Di\xfnm[ Y.]}, \bibinfo{author}{Xiao\xfnm[ X.]}.
\newblock \bibinfo{title}{The 10-m crop type maps in northeast china during 2017--2019}.
\newblock \bibinfo{journal}{Scientific data} \bibinfo{year}{2021};\bibinfo{volume}{8}(\bibinfo{number}{1}):\bibinfo{pages}{41}.
\bibitem[{Yuan et~al.(2020)Yuan, Shen, Li, Li, Li, Jiang, Xu, Tan, Yang, Wang et~al.}]{yuan2020deep}
\bibinfo{author}{Yuan\xfnm[ Q.]}, \bibinfo{author}{Shen\xfnm[ H.]}, \bibinfo{author}{Li\xfnm[ T.]}, \bibinfo{author}{Li\xfnm[ Z.]}, \bibinfo{author}{Li\xfnm[ S.]}, \bibinfo{author}{Jiang\xfnm[ Y.]}, \bibinfo{author}{Xu\xfnm[ H.]}, \bibinfo{author}{Tan\xfnm[ W.]}, \bibinfo{author}{Yang\xfnm[ Q.]}, \bibinfo{author}{Wang\xfnm[ J.]}, et~al.
\newblock \bibinfo{title}{Deep learning in environmental remote sensing: Achievements and challenges}.
\newblock \bibinfo{journal}{Remote sensing of Environment} \bibinfo{year}{2020};\bibinfo{volume}{241}:\bibinfo{pages}{111716}.
\bibitem[{Zhang et~al.(2023)Zhang, Chen, Li, Xiong and Lu}]{zhang2023sanet}
\bibinfo{author}{Zhang\xfnm[ B.]}, \bibinfo{author}{Chen\xfnm[ Y.]}, \bibinfo{author}{Li\xfnm[ Z.]}, \bibinfo{author}{Xiong\xfnm[ S.]}, \bibinfo{author}{Lu\xfnm[ X.]}.
\newblock \bibinfo{title}{Sanet: A self-attention network for agricultural hyperspectral image classification}.
\newblock \bibinfo{journal}{IEEE Transactions on Geoscience and Remote Sensing} \bibinfo{year}{2023};\bibinfo{volume}{62}:\bibinfo{pages}{1--15}.
\bibitem[{Zheng et~al.(2024)Zheng, Chen, Qian, Shi, Shu and Chen}]{zheng2024review}
\bibinfo{author}{Zheng\xfnm[ Y.]}, \bibinfo{author}{Chen\xfnm[ Y.]}, \bibinfo{author}{Qian\xfnm[ B.]}, \bibinfo{author}{Shi\xfnm[ X.]}, \bibinfo{author}{Shu\xfnm[ Y.]}, \bibinfo{author}{Chen\xfnm[ J.]}.
\newblock \bibinfo{title}{A review on edge large language models: Design, execution, and applications}.
\newblock \bibinfo{journal}{ACM Computing Surveys} \bibinfo{year}{2024};.
\bibitem[{Zhou et~al.(2020)Zhou, Sun, Luo, Zha and Zeng}]{zhou2020spatiotemporal}
\bibinfo{author}{Zhou\xfnm[ Y.]}, \bibinfo{author}{Sun\xfnm[ X.]}, \bibinfo{author}{Luo\xfnm[ C.]}, \bibinfo{author}{Zha\xfnm[ Z.J.]}, \bibinfo{author}{Zeng\xfnm[ W.]}.
\newblock \bibinfo{title}{Spatiotemporal fusion in 3d cnns: A probabilistic view}.
\newblock In: \bibinfo{booktitle}{proceedings of the IEEE/CVF conference on computer vision and pattern recognition}. \bibinfo{year}{2020}. p. \bibinfo{pages}{9829--9838}.
\bibitem[{Zhu et~al.(2017)Zhu, Tuia, Mou, Xia, Zhang, Xu and Fraundorfer}]{zhu2017deep}
\bibinfo{author}{Zhu\xfnm[ X.X.]}, \bibinfo{author}{Tuia\xfnm[ D.]}, \bibinfo{author}{Mou\xfnm[ L.]}, \bibinfo{author}{Xia\xfnm[ G.S.]}, \bibinfo{author}{Zhang\xfnm[ L.]}, \bibinfo{author}{Xu\xfnm[ F.]}, \bibinfo{author}{Fraundorfer\xfnm[ F.]}.
\newblock \bibinfo{title}{Deep learning in remote sensing: A comprehensive review and list of resources}.
\newblock \bibinfo{journal}{IEEE geoscience and remote sensing magazine} \bibinfo{year}{2017};\bibinfo{volume}{5}(\bibinfo{number}{4}):\bibinfo{pages}{8--36}.

\end{thebibliography}
\end{document}